\documentclass{article} % For LaTeX2e
\usepackage{iclr2021_conference,times}

\usepackage{amsmath,amsfonts,bm}

\def\eqref#1{equation~\ref{#1}}
\def\1{\bm{1}}

\DeclareMathAlphabet{\mathsfit}{\encodingdefault}{\sfdefault}{m}{sl}
\SetMathAlphabet{\mathsfit}{bold}{\encodingdefault}{\sfdefault}{bx}{n}

\usepackage{hyperref}
\usepackage{url}
\usepackage{booktabs}
\usepackage{wrapfig,lipsum,booktabs}
\usepackage{float}
\usepackage{subcaption}
\usepackage{graphicx}
\usepackage{wrapfig}
\usepackage{lineno}
\usepackage[final]{animate} % To compile with animations
\usepackage{hyperref}

\usepackage{bbm}
\usepackage{array}
\newcolumntype{L}[1]{>{\raggedright\let\newline\\\arraybackslash\hspace{0pt}}m{#1}}
\newcolumntype{C}[1]{>{\centering\let\newline\\\arraybackslash\hspace{0pt}}m{#1}}
\newcolumntype{R}[1]{>{\raggedleft\let\newline\\\arraybackslash\hspace{0pt}}m{#1}}

\newcommand{\ourApproach}{MoCoGAN-HD}
\title{A Good Image Generator Is What You Need for High-Resolution Video Synthesis}
\author{Yu Tian$^{1}$\thanks{Work done while at Snap Inc.}, Jian Ren$^2$, Menglei Chai$^2$, Kyle Olszewski$^2$, Xi Peng$^3$, \\
{\bf Dimitris N. Metaxas$^1$, Sergey Tulyakov$^2$} \\
$^1$Rutgers University, $^2$Snap Inc., $^3$University of Delaware\\
\texttt{\{yt219, dnm\}@cs.rutgers.edu}, \\\texttt{\{jren, mchai, kolszewski, stulyakov\}@snapchat.com}}

\iclrfinalcopy % Uncomment for camera-ready version, but NOT for submission.
\begin{document}

\maketitle
% \linenumbers
\begin{abstract}
Image and video synthesis are closely related areas aiming at generating content from noise. While rapid progress has been demonstrated in improving image-based models to handle large resolutions, high-quality renderings, and wide variations in image content, achieving comparable video generation results remains problematic.
We present a framework that leverages contemporary image generators to render high-resolution videos. We frame the video synthesis problem as discovering a trajectory in the latent space of a \emph{pre-trained} and \emph{fixed} image generator. Not only does such a framework render high-resolution videos, but it also is an order of magnitude more computationally efficient. We introduce a motion generator that discovers the desired trajectory, in which content and motion are disentangled.
With such a representation, our framework allows for a broad range of applications, including content and motion manipulation. Furthermore, we introduce a new task, which we call \emph{cross-domain video synthesis}, in which the image and motion generators are trained on disjoint datasets belonging to \emph{different} domains. This allows for generating moving objects for which the desired video data is not available. Extensive experiments on various datasets demonstrate the advantages of our methods over existing video generation techniques. Code will be released at \href{https://github.com/snap-research/MoCoGAN-HD}{\color{red}{\it{https://github.com/snap-research/MoCoGAN-HD}}}.

\end{abstract}

\section{Introduction}
Video synthesis seeks to generate a sequence of moving pictures from noise. While its closely related counterpart---image synthesis---has seen substantial advances in recent years, allowing for synthesizing at high resolutions~\citep{karras2017progressive}, rendering images often indistinguishable from real ones~\citep{karras2019style}, and supporting multiple classes of image content~\citep{zhang2019self}, contemporary improvements in the domain of video synthesis have been comparatively modest. Due to the statistical complexity of videos and larger model sizes, video synthesis produces relatively low-resolution videos, yet requires longer training times. For example, scaling the image generator of~\citet{brock2018large} to generate $256\times256$ videos requires a substantial computational budget\footnote{We estimate that the cost of training a model such as DVD-GAN~\citep{clark2019adversarial} once requires $>$ \$30K.}. Can we use a similar method to attain higher resolutions? We believe a different approach is needed.

There are two desired properties for generated videos: (i) high quality for each individual frame, and (ii) the frame sequence should be temporally consistent, \textit{i.e.} depicting the same content with plausible motion. Previous works~\citep{tulyakov2017mocogan,clark2019adversarial} attempt to achieve both goals with a single framework, making such methods computationally demanding when high resolution is desired.
We suggest a different perspective on this problem. We hypothesize that, given an image generator that has learned the distribution of video frames as independent images, a video can be represented as a sequence of latent codes from this generator. The problem of video synthesis can then be framed as discovering a latent trajectory that renders temporally consistent images. Hence, we demonstrate that (i) can be addressed by a \emph{pre-trained} and \emph{fixed} image generator, and (ii) can be achieved using the proposed framework to create appropriate image sequences.

%Besides, scaling existing video models to even higher resolutions requires disproportional increase of computation time making improvements in the domain unclear. 
% We term our framework as VIGOR as it performs VIdeo synthesis using image GeneratOR. To discover the sought latent trajectory, we introduce a latent motion generator, implemented via a recurrent neural network. The motion generator operates on the latent space to get motion representation, which is represented as a residual for latent content code. The disentangled motion residual, when combined with content code, correspond to a video frame.

To discover the appropriate latent trajectory, we introduce a motion generator, implemented via two recurrent neural networks, that operates on the initial content code to obtain the motion representation. 
We model motion as a residual between continuous latent codes that are passed to the image generator for individual frame generation.
Such a residual representation can also facilitate the disentangling of motion and content.
The motion generator is trained using the chosen image discriminator with contrastive loss to force the content to be temporally consistent, and a patch-based multi-scale video discriminator for learning motion patterns. Our framework supports contemporary image generators such as StyleGAN2~\citep{karras2019style} and BigGAN~\citep{brock2018large}.

%Our 
{We name our approach as \ourApproach{ }(Motion and Content decomposed GAN for High-Definition video synthesis) as it} features several major advantages over traditional video synthesis pipelines. First, it transcends the limited resolutions of existing techniques, allowing for the generation of high-quality videos at resolutions up to $1024\times1024$. Second, as we search for a latent trajectory in an image generator, our method is computationally more efficient, requiring an order of magnitude less training time than previous video-based works~\citep{clark2019adversarial}. Third, as the image generator is fixed, it can be trained on a separate high-quality image dataset.
Due to the disentangled representation of motion and content, our approach can learn motion from a video dataset and apply it to an image dataset, even in the case of two datasets belonging to \emph{different} domains. It thus unleashes the power of an image generator to synthesize high quality videos when a domain (\textit{e.g.}, dogs) contains many high-quality images but no corresponding high-quality videos (see Fig.~\ref{fig:cross-domain}). In this manner, our method can generate realistic videos of objects it has never seen moving during training (such as generating realistic pet face videos using motions extracted from images of talking people). We refer to this new video generation task as \emph{cross-domain video synthesis}. Finally, we quantitatively and qualitatively evaluate our approach, attaining state-of-the-art performance on each benchmark, and establish a challenging new baseline for video synthesis methods.

\section{Related Work}
\noindent\textbf{Video Synthesis}.
Approaches to image generation and translation using Generative Adversarial Networks (GANs)~\citep{goodfellow2014generative} have demonstrated the ability to synthesize high quality images~\citep{radford2015unsupervised,zhang2019self,brock2018large,donahue2019large,jin2021teachers}. Built upon image translation~\citep{isola2017image,wang2018high}, works on video-to-video translation~\citep{bansal2018recycle,wang2018vid2vid} are capable of converting an input video to a high-resolution output in another domain. However, the task of high-fidelity video generation, in the unconditional setting, is still a difficult and unresolved problem. Without the strong conditional inputs such as segmentation masks~\citep{wang2019fewshotvid2vid} or human poses~\citep{chan2019everybody,ren2020human} that are employed by video-to-video translation works, generating videos following the distribution of training video samples is challenging. Earlier works on GAN-based video modeling, including {MDPGAN~\citep{yushchenko2019markov}}, VGAN~\citep{vondrick2016generating}, TGAN~\citep{saito2017temporal}, MoCoGAN~\citep{tulyakov2017mocogan},  ProgressiveVGAN~\citep{acharya2018towards},
% \textcolor{blue}{MDPGAN~\citep{yushchenko2019markov},} 
TGANv2~\citep{TGAN2020} show promising results on low-resolution datasets. Recent efforts demonstrate the capacity to generate more realistic videos, but with significantly more computation~\citep{clark2019adversarial,Weissenborn2020Scaling}.
In this paper, we focus on generating realistic videos using manageable computational resources. 
{LDVDGAN~\citep{kahembwe2020lower} uses low dimensional discriminator to reduce model size and can generate videos with resolution up to $512\times512$, while we decrease training cost by utilizing a pre-trained image generator.}
% Our work does not rely on specific design and the computation efficiency is due to separately pre-training the image generator. Therefore, we are orthogonal to them and the combination with them would potentially lead to further efficiency.} 
The high-quality generation is achieved by using pre-trained image generators, while the motion trajectory is modeled within the latent space.
Additionally, learning motion in the latent space allows us to easily adapt the video generation model to the task of video prediction{~\citep{denton2017unsupervised}}, in which the starting frame is given~\citep{denton2018stochastic,zhao2018learning,walker2017pose,villegas17hierchvid,villegas17mcnet,babaeizadeh2017stochastic,hsieh2018learning,byeon2018contextvp}, by inverting the initial frame through the generator~\citep{abdal2020image2stylegan++}, instead of training an extra image encoder~\citep{tulyakov2017mocogan,zhang2020dtvnet}.

\noindent\textbf{Interpretable Latent Directions}.
The latent space of GANs is known to consist of semantically meaningful vectors for image manipulation. Both supervised methods, either using human annotations or pre-trained image classifiers~\citep{goetschalckx2019ganalyze,shen2020interpreting}, and unsupervised methods~\citep{jahanian2019steerability,plumerault2020controlling}, are able to find interpretable directions for image editing, such as supervising directions for image rotation or background removal~\citep{voynov2020unsupervised,shen2020closed}. We further consider the motion vectors in the latent space. By disentangling the motion trajectories in an unsupervised fashion, we are able to transfer the motion information from a video dataset to an image dataset in which no temporal information is available. 

\textbf{Contrastive Representation Learning}
is widely studied in unsupervised learning tasks~\citep{he2020momentum,chen2020simple,chen2020big,henaff2019data,lowe2019putting,oord2018representation,misra2020self}. Related inputs, such as images~\citep{wu2018unsupervised} or latent representations~\citep{hjelm2018learning}, which can vary  while training due to data augmentation, are forced to be close by minimizing differences in their representation during training. Recent work~\citep{park2020cut} applies noise-contrastive estimation~\citep{gutmann2010noise} to image generation tasks by learning the correspondence between image patches, achieving performance superior to that attained when using cycle-consistency constraints~\citep{zhu2017unpaired,yi2017dualgan}. 
On the other hand, we learn an image discriminator to create videos with coherent content by leveraging contrastive loss~\citep{hadsell2006dimensionality} along with an adversarial loss~\citep{goodfellow2014generative}.
\begin{figure}[t]
\begin{center}
\includegraphics[width=1\linewidth]{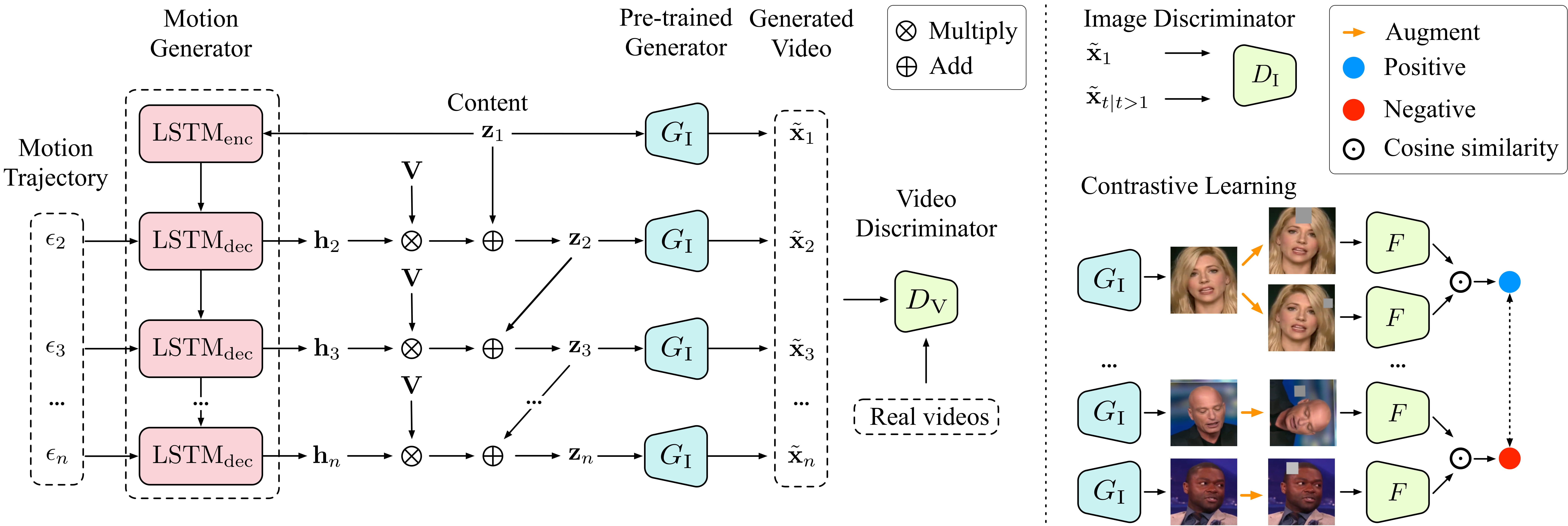}
\caption{\textbf{Left}: Given an initial latent code $\mathbf{z}_1$, a trajectory $\epsilon_t$, and a PCA basis $\mathbf{V}$, the motion generator $G_\mathrm{M}$ encodes $\mathbf{z}_1$ using $\mathrm{LSTM_{enc}}$ to get the initial hidden state and uses $\mathrm{LSTM_{dec}}$ to estimate hidden states for future frames. The image generator $G_\mathrm{I}$ synthesizes images using the predicted latent codes. The discriminator $D_\mathrm{V}$ is trained on both real and generated video sequences. 
% The mutual information between $\mathbf{h}_t$ and $\epsilon_t$ is maximized by $L_m$ so as to generate diverse motion patterns. 
\textbf{Right}: For each generated video, the first and subsequent frames are sent to an image discriminator $D_\mathrm{I}$. An encoder-like network $F$ 
%shares the weights with $D_\mathrm{I}$ and replaces the last layer with two fully-connected layers, 
calculates the features of synthesized images used to compute the contrastive loss $\mathcal{L}_\mathrm{contr}$ with positive (same image content, but different augmentation, shown in blue) and negative pairs (different image content and augmentation, shown in red).
}\label{fig:pipeline}
\end{center}
\end{figure}

\section{Method}
% Video generation is a much more challenging task than image generation, as a video generative model is required to additionally model temporal consistency and dynamic motion. Prior art suffers from two major problems: relative low frame resolution (128$\times$128) and video quality~\citep{tulyakov2017mocogan}, and high computational cost~\citep{clark2019adversarial} for training.
% Our framework is built on top of a \emph{pre-trained} image generator, since existing image synthesis works already demonstrate impressive results, even with relatively limited amounts of training data and computational resource~\citep{karras2020training,karras2020analyzing,zhao2020differentiable,zhao2020image}. 
In this section, we introduce our method for high-resolution video generation.
Our framework is built on top of a \emph{pre-trained} image generator~\citep{karras2020training,karras2020analyzing,zhao2020differentiable,zhao2020image}, which helps to generate high-quality image frames and boosts the training efficiency with manageable computational resources.
In addition, with the image generator fixed during training, we can disentangle video motion from image content, and enable video synthesis even when the image content and the video motion come from different domains.

More specifically, our inference framework includes a motion generator $G_\mathrm{M}$ and an image generator $G_\mathrm{I}$. $G_\mathrm{M}$ is implemented with two LSTM networks~\citep{hochreiter1997long} and predicts the latent motion \emph{trajectory} $\mathbf{Z} = \{\mathbf{z}_{1}, \mathbf{z}_{2}, \cdots, \mathbf{z}_{n}\}$, where $n$ is the number of frames in the synthesized video. The image generator $G_\mathrm{I}$ can thus synthesize each individual frame from the motion trajectory. The generated video sequence $\tilde{\mathbf{v}}$ is given by $\tilde{\mathbf{v}} = \{\tilde{\mathbf{x}}_1, \tilde{\mathbf{x}}_2, \cdots, \tilde{\mathbf{x}}_n\}$. For each synthesized frame $\tilde{\mathbf{x}}_t$, we have $\tilde{\mathbf{x}}_t = G_\mathrm{I}(\mathbf{z}_t)$  for $t = 1, 2, \cdots, n$. We also define the real video clip as $\mathbf{v} = \{\mathbf{x}_1, \mathbf{x}_2, \cdots, \mathbf{x}_n\}$ and the training video distribution as $p_v$.

To train the motion generator $G_\mathrm{M}$ to discover the desired motion trajectory, we apply a video discriminator to constrain the generated motion patterns to be similar to those of the training videos, and an image discriminator to force the frame content to be temporally consistent. Our framework is illustrated in Fig.~\ref{fig:pipeline}. We describe each component in more detail in the following sections.

% \textcolor{blue}{
% Consider the following thought experiment: suppose there is a perfect image generator $G$ in a specific domain (e.g. human faces). The word ``perfect'' here means for any given data $x$ in domain $\mathcal{X}$, $G$ has the ability to reproduce $x$ with a latent code $z\in \mathcal{Z}: x = G(z)$. With such an image generator $G$ at hand, any video $v = \{v_{1},v_{2},\cdots, v_{T}\} \in \mathcal{V}$, $v_{t} \in \mathcal{X}$ can be expressed as a \emph{trajectory} in $\mathcal{Z}$: $\{z_{1},z_{2},\cdots, z_{T}\}$, where $v_{t}=G(z_{t})$ for $t=\{1,2,\cdots, T\}$. Inspired by this thought experiment, as shown in Fig.~XX, we decouple the video generation into a two-step process: first, seek a ``perfect'' image generator, second, fix the image generator, train a RNN model to find the trajectory for video generation.}

% \textcolor{blue}{
% % \subsection{From Image to Video Generation}
% Given a well-trained image generator $G$, our video generation starts with a single random code $z\in \mathcal{Z}$. The first frame $v_{1}$ is created by the fixed generator $G: v_{1}=G(z)$. $v_{1}$ always has the state-of-the-art visual quality as it is created by the image generation process. Next we will show how to synthesize other frames as well as to control their visual quality and motion smoothness.
% }

\subsection{Motion Generator}\label{sec:motion_g}
The motion generator $G_\mathrm{M}$ predicts consecutive latent codes using an input code $\mathbf{z}_{1}\in \mathcal{{Z}}$, where the latent space $\mathcal{{Z}}$ is also shared by the image generator. 
For BigGAN~\citep{brock2018large}, we sample $\mathbf{z}_{1}$ from the normal distribution $p_{z}$.
% \textcolor{red}{Adding BigGAN depending on experiments.}
For StyleGAN2~\citep{karras2020analyzing}, $p_z$ is the distribution  
% use the latent space $\mathcal{W}$ 
after the multi-layer perceptron (MLP), %$M$
as the latent codes within this distribution can be semantically disentangled better than when using the normal distribution~\citep{shen2020interpreting,zhu2020indomain}.
%  image semantics
% To facilitate the presentation, we use $\mathcal{Z}$ to represent the image latent space.

% Given the first latent code $z_{1}\in \mathcal{Z}$, our goal is to generate consecutive latent codes $z_{2}, z_{3}, \cdots, z_{T}$ such that the induced video frames $v_{t}=G(z_{t}), t=2,3,\cdots,T$ share the same \emph{content} with $v_{1}$, while the whole sequence $v=\{v_{1},v_{2},\cdots,v_{T}\}$ represents some motion patterns of the real video dataset. Given different choices of generator, we choose different latent distribution. $\mathcal{W}$ for StyleGAN2~\citep{karras2020analyzing} while $\mathcal{Z}$ for BigGAN~\citep{brock2018large}

Formally, $G_\mathrm{M}$ includes an LSTM encoder $\mathrm{LSTM}_\text{enc}$, which encodes $\mathbf{z}_1$ to get the initial hidden state, and a LSTM decoder $\mathrm{LSTM}_\text{dec}$, which  estimates $n - 1$ continuous hidden states recursively:
\begin{equation}\label{eqn:lstm}
\begin{gathered}
\mathbf{h}_{1}, \mathbf{c}_{1} = \mathrm{LSTM}_\text{enc}(\mathbf{z}_{1}), \\
\mathbf{h}_{t}, \mathbf{c}_{t} = \mathrm{LSTM}_\text{dec}(\epsilon_{t}, (\mathbf{h}_{t-1}, \mathbf{c}_{t-1})), ~~~ t = 2, 3, \cdots, n,
\end{gathered}
\end{equation}
where $\mathbf{h}$ and $\mathbf{c}$ denote the hidden state and cell state respectively, and $\epsilon_{t}$ is a noise vector sampled from the normal distribution to model the motion diversity at timestamp $t$. % $\mathcal{E}$ that $\mathcal{E} \sim \mathcal{N}(\textbf{0},\textbf{I})$. % We then discuss how to disentangle motion and content from $h_t$,  and use $e_t$ for generation diverse motion with the same content.

{\bf Motion Disentanglement.}
Prior work~\citep{tulyakov2017mocogan} applies $\mathbf{h}_{t}$ as the motion code for the frame to be generated, while the content code is fixed for all frames. However, such a design requires a recurrent network to estimate the motion while preserving consistent content from the latent vector, which is difficult to learn in practice. Instead, we propose to use a sequence of motion residuals
%disentangling the motion from content via a residual implementation 
for estimating the motion trajectory. 
%Inspired by recent work on learning useful directions in the latent distribution for image editing~\citep{shen2020closed,harkonen2020ganspace}, 
{Specifically, we model the motion residual as the linear combination of a set of interpretable directions in the latent space~\citep{shen2020closed,harkonen2020ganspace}.}
We first conduct principal component analysis (PCA) on $m$ randomly sampled latent vectors from $\mathcal{Z}$ to get the basis $\mathbf{V}$. Then, we estimate the motion direction from the previous frame $\mathbf{z}_{t-1}$ to the current frame $\mathbf{z}_t$ by using $\mathbf{h}_t$ and $\mathbf{V}$ as follows:
\begin{equation}\label{eqn:residual}
    \mathbf{z}_{t} = \mathbf{z}_{t-1} + \lambda\cdot  \mathbf{h}_{t} \cdot \mathbf{V}, ~~~ t = 2, 3, \cdots, n,
\end{equation}
where the hidden state $\mathbf{h}_{t}\in [-1,1]$, and $\lambda$ controls the step given by the residual. With Eqn.~\ref{eqn:lstm} and Eqn.~\ref{eqn:residual}, we have $G_\mathrm{M}(\mathbf{z}_1) = \{\mathbf{z}_{1}, \mathbf{z}_{2}, \cdots, \mathbf{z}_{n}\}$, and the generated video $\tilde{\mathbf{v}}$ is given as $\tilde{\mathbf{v}} = G_\mathrm{I}(G_\mathrm{M}(\mathbf{z}_1))$.

{\bf Motion Diversity.} In Eqn.~\ref{eqn:lstm}, we introduce a noise vector $\epsilon_t$ to control the diversity of motion. However, we observe that the LSTM decoder tends to neglect the $\epsilon_t$, resulting in \textit{motion mode collapse}, meaning that $G_\mathrm{M}$ cannot capture the diverse motion patterns from training videos and generate distinct videos from one initial latent code with similar motion patterns for different sequences of noise vectors. To alleviate this issue, we introduce a mutual information loss $\mathcal{L}_\mathrm{m}$ to maximize the \textit{mutual information} between the hidden vector $\mathbf{h}_t$ and the noise vector $\epsilon_t$. With $\mathrm{sim}(\mathbf{u},\mathbf{v})=\mathbf{u}^{T}\mathbf{v}/\left \| \mathbf{u} \right \|\left \| \mathbf{v} \right \|$ denoting the cosine similarity between vectors $\mathbf{u}$ and $\mathbf{v}$, we define $\mathcal{L}_\mathrm{m}$ as follows:
\begin{equation}\label{eqn:mut}
\begin{gathered}
 \mathcal{L}_\mathrm{m} = \frac{1}{n-1} \sum_{t=2}^{n} \mathrm{sim}(H(\mathbf{h}_{t}),\epsilon_{t}), \\
\end{gathered}
\end{equation}
where $H$ is a 2-layer MLP that serves as a mapping function.  

% In image generation, mode collapse indicates the generator can not capture most modes of the training dataset, and instead generates repetitive images~\citep{goodfellow2016nips}. In video generation, we observed another mode collapse: motion mode collapse, where the generator can only synthesis videos with highly correlated motion. As in Eq.~\ref{lstm-dec}, $e_{t}$ controls the randomness of $h_t$. We discovered that without any supervision, $\mathrm{LSTM}_{dec}$ tends to ignore $e_{t}$, and generates $h_{t}$ with $h_{t-1}$ and $c_{t-1}$ directly. As a consequence, the whole model produces videos with similar motion.

% To remedy this problem, in Eq.~\ref{lstm-dec}, $h_{t}$ should contain the information from $e_{t}$. To this end, we propose to maximize the \emph{mutual information} between $h_{t}$ and $e_{t}$. Specifically, $h_{t}$ should has the ability to rediscover $e_t$, therefore, we introduce the following mutual information loss:
% \begin{equation}
%     L_{mut} = \mathrm{sim}(\mathrm{MLP}(h_{t}),e_{t}),
% \end{equation}
% where $\mathrm{sim}(u,v)=u^{T}v/\left \| u \right \|\left \| v \right \|$ denote the cosine similarity. In experiments (Sec.XXX) we show this improves the diversity of generated videos.

{\bf Learning.}
To learn the appropriate parameters for the motion generator $G_\mathrm{M}$, we apply a multi-scale video discriminator $D_\mathrm{V}$ to tell whether a video sequence is real or synthesized.  The discriminator is based on the architecture of PatchGAN~\citep{isola2017image}. However, we use 3D convolutional layers in $D_\mathrm{V}$, as they can model temporal dynamics better than 2D convolutional layers. We divide input video sequence into small %$N\times N\times N$
3D patches, and classify each patch as real or fake. The local responses for the input sequence are averaged to produce the final output. Additionally, each frame in the input video sequence is conditioned on the first frame, as it falls into the distribution of the pre-trained image generator, for more stable training. We thus optimize the following  adversarial loss to learn $G_\mathrm{M}$ and $D_\mathrm{V}$:
% \citep{zhang2019self,brock2018large} 
% \begin{equation}\label{eqn:dv}
% \begin{gathered}
%     L_{D_v}=-{\mathbb{E}_{{v\sim p_v}}}\left [ \mathrm{min}(0,-1+D_v(v)) \right ] - {\mathbb{E}_{{z_{1}\sim p_z}}}\left [ \mathrm{min}(0,-1-D_v(G_I({G_m(z_1)}))) \right ], \\
%     L_{GV}=-{\mathbb{E}_{{z_{1}\sim p(z)}}}(D_v(G_I({G}_m(z_1)))).
% \end{gathered}
% \end{equation}
\begin{equation}\label{eqn:dv}
    \mathcal{L}_{D_\mathrm{V}}= {\mathbb{E}_{{\mathbf{v}\sim p_v}}}\left [\log D_\mathrm{v}(\mathbf{v})  \right ] + {\mathbb{E}_{{\mathbf{z}_{1}\sim p_{z}}}}\left[ \log (1 - D_\mathrm{V}(G_\mathrm{I}({G}_\mathrm{M}(\mathbf{z}_1)))) \right ].
\end{equation}

% So far we have 1) The well-trained image generator which guarantees to synthesis photo-realistic first frame; 2) Contrastive 2D discriminator which matches later frames to the first one; 3) An incremental RNN model which facilities model training. Therefore, we hope the 3D discriminator can focus more on the motion. Isola \emph{et al}~\citep{isola2017image} have proposed a 2D Markovian discriminator (PatchGAN), where they \textcolor{blue}{restricting the discriminator to only model high-frequency structure. In order to model
% high-frequencies, it is sufficient to restrict our attention to the structure in local image patches. Therefore, we design a discriminator architecture – which we term a PatchGAN – that only penalizes structure at the scale of patches. This discriminator tries to classify if each $N\times N$ patch in an image
% is real or fake. We run this discriminator convolutionally across the image, averaging all responses to provide the ultimate output of D.} Our 3D discriminator extends their ideas, where we divide the input video into $N\times N\times N$ patches, and use 3D convolution to get every local responses.

\subsection{Contrastive Image Discriminator}\label{sec:con_d}
% \textcolor{red}{to do}
As our image generator is pre-trained, we may use an image generator that is trained on a given domain, \textit{e.g.} images of animal faces~\citep{choi2020starganv2}, and learn the motion generator parameters using videos from a different domain, such as videos of human facial expressions~\citep{Nagrani17}. 
With Eqn.~\ref{eqn:dv} alone, however, we lack the ability to explicitly constrain the generated images $\tilde{\mathbf{x}}_{t|t>1}$ to possess similar quality and content as the first image $\tilde{\mathbf{x}}_{1}$, which is sampled from the image space of the image generator and thus has high fidelity.
Hence, we introduce a contrastive image discriminator $D_\mathrm{I}$, which is illustrated in Fig.~\ref{fig:pipeline}, to match both image quality and content between $\tilde{\mathbf{x}}_1$ and $\tilde{\mathbf{x}}_{t|t>1}$.

{\bf Quality Matching.} To increase the perceptual quality, we train $D_\mathrm{I}$ and $G_\mathrm{M}$ adversarially by forwarding $\tilde{\mathbf{x}}_t$ into the discriminator $D_\mathrm{I}$ and using $\tilde{\mathbf{x}}_1$ as real sample and $\tilde{\mathbf{x}}_{t|t>1}$ as the fake sample.
% \begin{equation}\label{eqn:dc}
% \small
% \begin{gathered}
%     L_{D_c}=-{\mathbb{E}_{{z_{1}\sim p_z}}}\left [ \mathrm{min}(0,-1+D_c(G_I(z_{1}))) \right ] - {\mathbb{E}_{{z_{i}\sim G_m(z_{1})}| i>1}}\left [ \mathrm{min}(0,-1-D_c(G_I(z_{i}))) \right ], \\
%     % L_{R_c}=-{\mathbb{E}_{{z_{1}\sim p(z),z_{i}\sim R(z_{1})}}}(D(G(z_{i}))).
%     L_{GC}=-{\mathbb{E}_{{z_{1}\sim p_z,z_{i}\sim G_m(z_{1})}| i>1}}(D_c(G_I(z_{i}))).
% \end{gathered}
% \end{equation}
\begin{equation}\label{eqn:dc}
    \mathcal{L}_{D_\mathrm{I}}= {\mathbb{E}_{{\mathbf{z}_{1}\sim p_z}}}\left [\log D_\mathrm{I}(G_\mathrm{I}(\mathbf{z}_1))  \right ] + {\mathbb{E}_{{\mathbf{z}_{1}\sim p_z, \mathbf{z}_{t}\sim G_\mathrm{M}(\mathbf{z}_{1})}| t>1}}\left[ \log (1 - D_\mathrm{I}(G_\mathrm{I}(\mathbf{z}_t))) \right ].
\end{equation}

{\bf Content Matching.} To learn content similarity between frames within a video, we use the image discriminator as a feature extractor and train it with a form of contrastive loss known as InfoNCE~\citep{oord2018representation}. The goal is that pairs of images with the same content should be close together in embedding space, while images containing different content should be far apart.

% Inspired by recent success in contrastive learning~\citep{chen2020simple,chen2020big,he2020momentum},
% , which we refer as the ``positive'' samples for the query.
% \textcolor{blue}{Following~\citep{chen2020simple,chen2020big,he2020momentum}, 
% Each augmented pair ($v_{t}^{(ia)}$, $v_{t}^{(ib)}$) is defined as positive pair, and 
%Other $2N-2$ augmented examples $ v_{t}^{(i\cdot)} v_{t}^{(j\cdot)}$ within this minibatch are defined as negative pairs. 
% \textcolor{blue}{To Do: Given a minibatch of $N$ videos $\{{v}^{(1)}, {v}^{(2)}, \cdots, {v}^{(N)}\}$, we randomly sample one frame $t$ from each video: $\{{x}^{(1)}_{t},{x}^{(2)}_{t},\cdots,{x}^{(N)}_{t}\}$,
% and make two random augmentation copies (${x}_{t}^{(ia)}$, ${x}_{k}^{(ib)}$) for each frame ${x}_{t}^{(i)}$, resulting in $2N$ samples. (${x}_{t}^{(ia)}$, ${x}_{t}^{(ib)}$) are positive pairs as they share the same content. (${x}_{t}^{(i\cdot)}$, ${x}_{t}^{(j\cdot)}$) are all negative pairs for $i\neq j$.
% }

 Given a minibatch of $N$ generated videos $\{\tilde{\mathbf{v}}^{(1)}, \tilde{\mathbf{v}}^{(2)}, \cdots, \tilde{\mathbf{v}}^{(N)}\}$, we randomly sample one frame $t$ from each video: $\{\tilde{\mathbf{x}}^{(1)}_{t},\tilde{\mathbf{x}}^{(2)}_{t},\cdots, \tilde{\mathbf{x}}^{(N)}_{t}\}$,
% which we refer as ``query'', 
and make two randomly augmented versions ($\tilde{\mathbf{x}}_{t}^{(ia)}$, $\tilde{\mathbf{x}}_{t}^{(ib)}$) for each frame $\tilde{\mathbf{x}}_{t}^{(i)}$, resulting in $2N$ samples. ($\tilde{\mathbf{x}}_{t}^{(ia)}$, $\tilde{\mathbf{x}}_{t}^{(ib)}$) are positive pairs, as they share the same content. ($\tilde{\mathbf{x}}_{t}^{(i\cdot)}$, $\tilde{\mathbf{x}}_{t}^{(j\cdot)}$) are all negative pairs for $i\neq j$.

Let $F$ be an encoder network, which shares the same weights and architecture of $D_\mathrm{I}$, but excluding the last layer of $D_\mathrm{I}$ and including a 2-layer MLP as a projection head %{~\citep{chen2020simple}} %(\textcolor{red}{but with a projection head}), 
that produces the representation of the input images. We have a contrastive loss function $\mathcal{L}_\mathrm{contr}$, which is the cross-entropy computed across $2N$ augmentations as follows:
\begin{equation}\label{eqn:con}
\small
\begin{split}
    \mathcal{L}_\mathrm{contr} = -\sum_{i=1}^{N}\sum_{\alpha=a}^{b} \log \frac{\exp(\mathrm{sim}(F (\tilde{\mathbf{x}}_{t}^{(ia)}),F(\tilde{\mathbf{x}}_{t}^{(ib)}))/\tau)}{\sum_{j=1}^{N}\sum_{\beta=a}^{b}\mathbbm{1}_{[j\neq i]}(\exp(\mathrm{sim}(F(\tilde{\mathbf{x}}_{t}^{(i\alpha)}),F(\tilde{\mathbf{x}}_{t}^{(j\beta)}))/\tau)}, %\\
    %-\sum_{i=1}^{N} \log \frac{\exp(\mathrm{sim}(F(\tilde{x}_{t}^{(ib)}),F(\tilde{x}_{t}^{(ia)}))/\tau)}{\sum_{j=1}^{N}\mathbbm{1}_{[j\neq i]}(\exp(\mathrm{sim}(F(\tilde{x}_{t}^{(ib)}),F(\tilde{x}_{t}^{(ja)}))/\tau)+\exp(\mathrm{sim}(F(\tilde{x}_{t}^{(ib)}),F(\tilde{x}_{t}^{(jb)}))/\tau))} ,
\end{split}
\end{equation}
% }
where $\text{sim}(\cdot, \cdot)$ is the cosine similarity function defined in Eqn.~\ref{eqn:mut}, $\mathbbm{1}_{[j\neq i]}\in \{0,1\}$ is equal to 1
iff $j\neq i$, and $\tau$ is a temperature parameter empirically set to $0.07$. We use a momentum decoder mechanism similar to that of MoCo~\citep{he2020momentum} by maintaining a memory bank to delete the oldest negative pairs and update the new negative pairs. We apply augmentation methods including translation, color jittering, and cutout~\citep{devries2017improved} on synthesized images. With the positive and negative pairs generated on-the-fly during training, the discriminator can effectively focus on the content of the input samples. %To achieve more stable training,
% \textcolor{blue}{To make the content inside a video consistent,}

The choice of positive pairs in Eqn.~\ref{eqn:con} is specifically designed for cross-domain video synthesis, as videos of arbitrary content from the image domain is not available. In the case that images and videos are from the same domain, the positive and negative pairs are easier to obtain. We randomly select and augment two frames from a real video to create positive pairs sharing the same content, while the negative pairs contain augmented images from different real videos. 

Aside from $\mathcal{L}_\mathrm{contr}$, we also adopt the feature matching loss~\citep{wang2018high} $\mathcal{L}_\mathrm{f}$ between the generated first frame and other frames by changing the $L_1$ regularization to cosine similarity.

% \textcolor{blue}{When the image generator and the video dataset are in the same domain, we modify the contrastive image discriminator to leverage real video frames as the positive pairs. See appendix for more details.}

{\bf Full Objective.}
The overall loss function for training motion generator, video discriminator, and image discriminator is thus defined as:
% \begin{equation}\label{eqn:full}
%     \min\limits_{G_\mathrm{M}}(\max\limits_{D_\mathrm{V}} L_{D_\mathrm{V}}(G_\mathrm{M}, D_\mathrm{V}) + \max\limits_{D_\mathrm{I}} L_{D_\mathrm{I}}(G_\mathrm{M}, D_\mathrm{I})) + \lambda_{m} L_m + \lambda_{con} L_{cont} + \lambda_{f} L_f,
% \end{equation}
% \begin{equation}\label{eqn:full}
%     \min\limits_{G_\mathrm{M}}(\max\limits_{D_\mathrm{V}} \mathcal{L}_{D_\mathrm{V}}(G_\mathrm{M}, D_\mathrm{V}) + \max\limits_{D_\mathrm{I}} \mathcal{L}_{D_\mathrm{I}}(G_\mathrm{M}, D_\mathrm{I}) + \lambda_\mathrm{m} \mathcal{L}_\mathrm{m} + \lambda_\mathrm{f} \mathcal{L}_\mathrm{f}) + \min\limits_{D_\mathrm{I}}(\lambda_\mathrm{contr} \mathcal{L}_\mathrm{contr})
% \end{equation}
\begin{equation}\label{eqn:full}
    \min\limits_{G_\mathrm{M}}(\max\limits_{D_\mathrm{V}} \mathcal{L}_{D_\mathrm{V}} + \max\limits_{D_\mathrm{I}} \mathcal{L}_{D_\mathrm{I}})
    + \max\limits_{G_\mathrm{M}}(\lambda_\mathrm{m}\mathcal{L}_\mathrm{m} + \lambda_\mathrm{f} \mathcal{L}_\mathrm{f}) + \min\limits_{D_\mathrm{I}}(\lambda_\mathrm{contr} \mathcal{L}_\mathrm{contr})
\end{equation}
where $\lambda_\mathrm{m}$, $\lambda_\mathrm{contr}$, and $\lambda_\mathrm{f}$ are hyperparameters to balance losses. 

\section{Experiments}
In this section, we evaluate the proposed approach on several benchmark datasets for video generation. We also demonstrate cross-domain video synthesis for various image and video datasets.
% and content-motion disentanglement and diverse video generation.

\subsection{Video Generation}\label{sec:video_generation}
We conduct experiments on three datasets including UCF-101~\citep{soomro2012ucf101}, FaceForensics~\citep{rossler2018faceforensics}, and Sky Time-lapse~\citep{xiong2018learning} for unconditional video synthesis. We use StyleGAN2 as the image generator.
Training details can be found in Appx.~\ref{app:experiments}.

\begin{table}[h]
\begin{minipage}[t] {.45\linewidth}
\centering
    \caption{\label{tab:ucf-101} {IS and FVD} on UCF-101.}
    \begin{tabular}{c|cc}
    \toprule 
    Method & IS ($\uparrow$) & FVD ($\downarrow$)  \\ \midrule
    VGAN & 8.31 $\pm$ .09 & - \\
    TGAN & 11.85 $\pm$ .07 & - \\
    MoCoGAN & 12.42 $\pm$ .07 & - \\
    ProgressiveVGAN &  14.56 $\pm$ .05 & - \\
    {LDVD-GAN} & {22.91 $\pm$ .19} & - \\
    TGANv2 & 26.60 $\pm$ .47 & 1209 $\pm$ 28 
    %1209.45957+-27.58755859549915
    \\
    DVD-GAN & 27.38 $\pm$ .53 & - \\ \midrule
    % Ours (128$\times$128) & \textbf{xxx $\pm$ \textcolor{red}{xx}} & \textcolor{red}{xxx} \\
    Ours  & \textbf{33.95 $\pm$ .25} & \textbf{700 $\pm$ 24}
    %700.166073+-23.881134348074028
    \\
    \bottomrule
    \end{tabular}
\end{minipage}\hfill
\begin{minipage}[t] {.46\linewidth}
\centering
    \caption{\label{tab:face} FVD, ACD, and Human Preference on FaceForensics.}
    \centering
    \begin{tabular}{c|cc}
    \toprule 
    Method & FVD ($\downarrow$)  & ACD ($\downarrow$)   \\ \midrule
    GT & 9.02 & 0.2935 \\ \midrule
    TGANv2  & 58.03
    %58.03326  
    & 0.4914  \\
    Ours & \textbf{53.26} %53.261326
    &  \textbf{0.3300}
    %0.32998104676265333
    \\
    \bottomrule
    \end{tabular}
    \begin{tabular}{c|c}
    \toprule 
    Method & Human Preference (\%)  \\ \midrule
     Ours / TGANv2   &  \textbf{73.6} / 26.4 \\
    % TGANv2 / GT  & \textcolor{red}{xxx} \\
    % Ours / GT  & 0.197 / 0.803 \\
    \bottomrule
\end{tabular}
\end{minipage}
\end{table}
% \subsubsection{UCF-101}

\textbf{UCF-101} is widely used in video generation. The dataset includes $13,320$ videos of $101$ sport categories. The resolution of each video is $320\times240$. To process the data, we crop a rectangle with size of $240\times240$ from each frame in a video and resize it to $256\times256$. 
We train the motion generator to predict $16$ frames. For evaluation, we report Inception Score (IS)~\citep{TGAN2020} on $10,000$ generated videos and Fr$\mathrm{\acute{e}}$chet Video Distance (FVD)~\citep{unterthiner2018towards} on $2,048$ videos. 
The classifier used to calculate IS is a C3D network~\citep{tran2015learning} that is trained on the Sports-1M dataset~\citep{karpathy2014large} and fine-tuned on UCF-101, which is the same model used in previous works~\citep{TGAN2020,clark2019adversarial}.

% The implementation of IS is the same as previous works~\citep{TGAN2020,clark2019adversarial} for a fair comparison.

% \textcolor{blue}{UCF101 is a common video dataset that contains 13,320 videos with 320$\times$240 pixels and 101 different sport categories. In the experiments, we randomly extracted 16 frames from the training dataset, cropped a rectangle with 240$\times$240 pixels from the center, resized it to 256$\times$256 pixels.} We train an unconditional image generator with 256$\times$256 resolution with StyleGAN2 model. \textcolor{red}{this goes to appendix: which has image FID 45.63} For video evaluation, we report Inception Score (IS)~\citep{TGAN2020} on 10,000 generations and Fr$\mathrm{\acute{e}}$chet Video Distance (FVD)~\citep{unterthiner2018towards} on 2,048 generations.

The quantitative results are shown in Tab.~\ref{tab:ucf-101}. Our method achieves state-of-the-art results for both IS and FVD, and outperforms existing works by a large margin. 
Interestingly, this result indicates that a well-trained image generator has learned to represent rich motion patterns, and therefore can be used to synthesize high-quality videos when used with a well-trained motion generator.

% \begin{table}
% \centering
% \begin{tabular}{lcccc}
% \hline 
% Method & first & Ours & TGANv2 & DVDGAN  \\ \hline
% ICP & 27.60 & {\bf 34.23} & 26.60 $\pm$ .47 & 27.38 $\pm$ .53\\
% \hline
% \end{tabular}
% \caption{\label{ucf-101} Inception Score on UCF-101 dataset.}
% \end{table}
% \begin{wraptable}{L}{0.45\linewidth}
% \centering
%     \caption{\label{ablation} Ablation Analysis on UCF-101.}
%     \begin{tabular}{c|cc}
%     \toprule 
%     Method & IS ($\uparrow$)  & FVD ($\downarrow$) \\ \midrule
%     baseline \\ (RNN + multiscale 3D)  &  &  \\
%     + PCA residual  &  &  \\
%     + contrastive &  &  \\
%     + mutual information \\(image from FFHQ)  &  &  \\
%     \bottomrule
%     \end{tabular} 
% \end{wraptable}

% \begin{figure}[h]
% \begin{center}
% \includegraphics[width=1\linewidth]{faceforensics.jpg}
% \caption{Example generated videos for the model trained on FaceForensics. We can generate  natural and photo-realistic videos with various motion patterns, such as eye blink and talking. Four examples show frames $2$, $7$, $11$, and $16$. }\label{fig:forensics}
% \end{center}
% \end{figure}

\textbf{FaceForensics} is a dataset containing news videos featuring various reporters. We use all the images from $704$ training videos, with a resolution of $256\times256$, to learn an image generator, and sequences of $16$ consecutive frames to train motion generator. Note that our network can generate even longer continuous sequences, \textit{e.g.} $64$ frames (Fig.~\ref{fig:face_64_appendix} in Appx.), though only $16$ frames are used for training. 

\begin{wrapfigure}{R}{0.5\linewidth}
% \begin{figure}
 \begin{center}
    \includegraphics[width=1\linewidth]{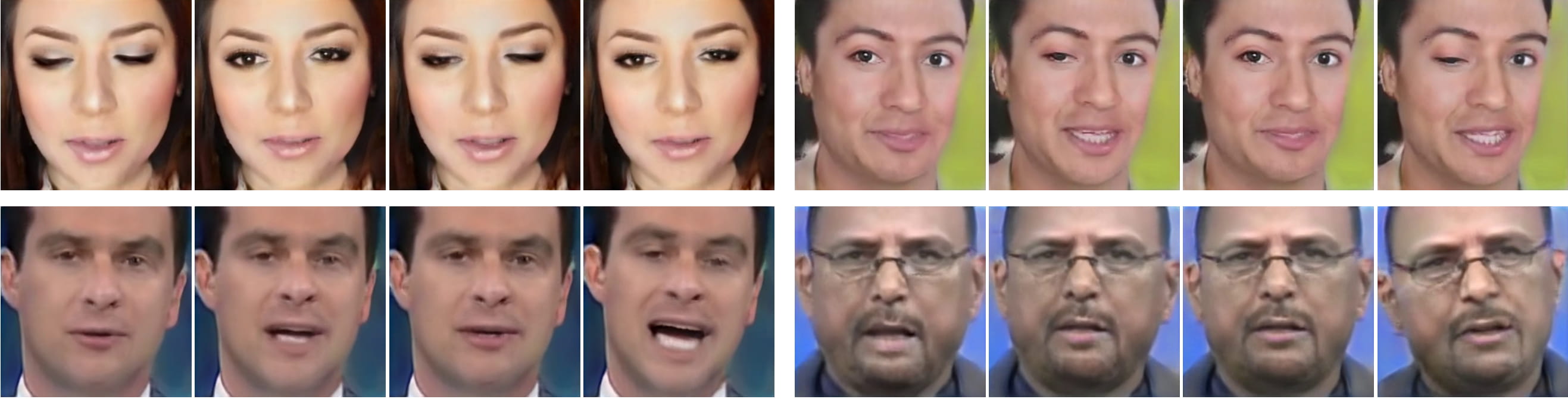}
  \end{center}
  \caption{Example generated videos from a model trained on FaceForensics. We can generate  natural and photo-realistic videos with various motion patterns, such as eye blink and talking. Four examples show frames $2$, $7$, $11$, and $16$. }\label{fig:forensics}
% \end{figure}
\end{wrapfigure}

We show the FVD between generated and real video clips ($16$ frames in length) for different methods in Tab.~\ref{tab:face}. Additionally, we use the Average Content Distance (ACD) from MoCoGAN~\citep{tulyakov2017mocogan} to evaluate the identity consistency for these human face videos. We calculate ACD values over $256$ videos. We also report the two metrics for ground truth (GT) videos. To get FVD of GT videos, we randomly sample two groups of real videos and compute the score.
Our method achieves better results than TGANv2~\citep{TGAN2020}. Both methods have low FVD values, and can generate complex motion patterns close to the real data. However, the much lower ACD value of our approach, which is close to GT, demonstrates that the videos it synthesizes have much better identity consistency than the videos from TGANv2. Qualitative examples in Fig.~\ref{fig:forensics} illustrate different motions patterns learned from the dataset.
Furthermore, we perform perceptual experiments using Amazon Mechanical Turk (AMT) by presenting a pair of videos from the two methods to users and asking them to select a more realistic one. Results in Tab.~\ref{tab:face} indicate our method outperforms TGANv2 in 73.6\% of the comparisons.

\textbf{Sky Time-Lapse} is a video dataset consisting of dynamic sky scenes, such as moving clouds. The number of video clips for training and testing is $35,392$ and $2,815$, respectively. 
We resize images to $128\times128$ and train the model to generate $16$ frames.
We compare our methods with the two recent approaches of MDGAN~\citep{xiong2018learning} and DTVNet~\citep{zhang2020dtvnet}, which are specifically designed for this dataset. In Tab.~\ref{tab:sky}, we report the FVD for all three methods. It is clear that our approach significantly outperforms the others. Example sequences are shown in Fig.~\ref{fig:sky}. 

Following DTVNet~\citep{zhang2020dtvnet}, we evaluate the proposed model for the task of \textit{video prediction}. We use the Peak Signal-to-Noise Ratio (PSNR) and Structural Similarity (SSIM)~\citep{wang2004image} as evaluation metrics to measure the frame quality at the pixel level and the structural similarity between synthesized and real video frames.
Evaluation is performed on the testing set. We select the first frame $\mathbf{x}_1$ from each video clip and project it to the latent space of the image generator~\citep{abdal2020image2stylegan++} to get $\hat{\mathbf{z}}_1$. 
We use $\hat{\mathbf{z}}_1$ as the starting latent code for motion generator to get $16$ latent codes, and interpolate them to get $32$ latent codes to synthesize a video sequence, where the first frame is given by $G_\mathrm{I}(\hat{\mathbf{z}}_1)$.
For a fair comparison, we also use $G_\mathrm{I}(\hat{\mathbf{z}}_1)$ as the starting frame for MDGAN and DTVNet to calculate the metrics with ground truth videos. In addition, we calculate the PSNR and SSIM between $\mathbf{x}_1$ and $G_\mathrm{I}(\hat{\mathbf{z}}_1)$ as the upper bound for all methods, which we denote as \textit{Up-B}.
Tab.~\ref{tab:sky} shows the video prediction results, which demonstrate that our method's performance is superior to those of MDGAN and DTVNet. Interestingly, by simply interpolating the motion trajectory, we can easily generate longer video sequence, \textit{e.g.} from $16$ to $32$ frames, while retaining high quality.

\begin{minipage}[h]{\linewidth}
\begin{minipage}{0.45\linewidth}
    \centering
    \captionof{table}{Evaluation on Sky Time-lapse for video synthesis and prediction.}\label{tab:sky} 
    \begin{tabular}{c|ccc}
    \toprule
    Method & FVD ($\downarrow$) & PSNR ($\uparrow$)  & SSIM ($\uparrow$)   \\ \midrule
    Up-B & - & 25.367 & 0.781  \\ \midrule
    MDGAN  & 840.95 & 13.840 & 0.581 \\
    DTVNet & 451.14 & 21.953 & 0.531 \\
    Ours & \textbf{77.77} & \textbf{22.286} &\textbf{0.688} \\
    \bottomrule
    \end{tabular} 
    \end{minipage} \hfill
  \begin{minipage}{0.45\linewidth}
    \centering
    \includegraphics[width=\linewidth]{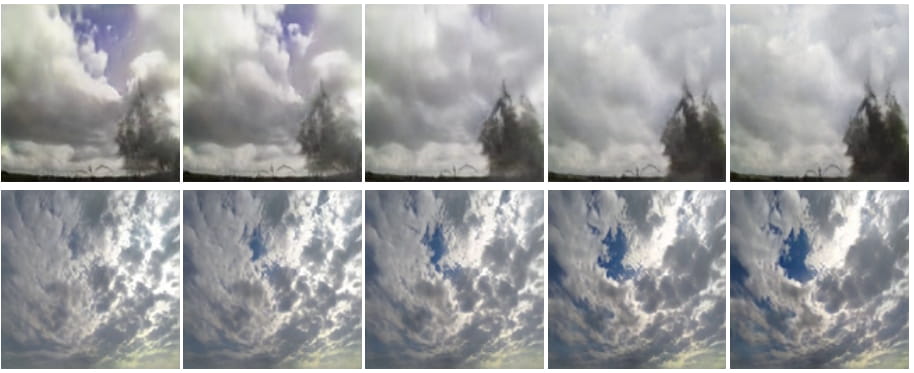}
    \setlength{\tabcolsep}{0\linewidth}
\begin{tabular}{C{0.2\linewidth}C{0.2\linewidth}C{0.2\linewidth}C{0.2\linewidth}C{0.2\linewidth}}
\small$t=2$&\small$t=6$&\small$t=10$&\small$t=14$&\small$t=16$
\end{tabular}
    \captionof{figure}{Sample generated frames at several time steps ($t$) for the Sky Time-lapse dataset.}\label{fig:sky}
  \end{minipage}
\end{minipage}

\begin{figure}[h]
\begin{center}
\includegraphics[width=1\linewidth]{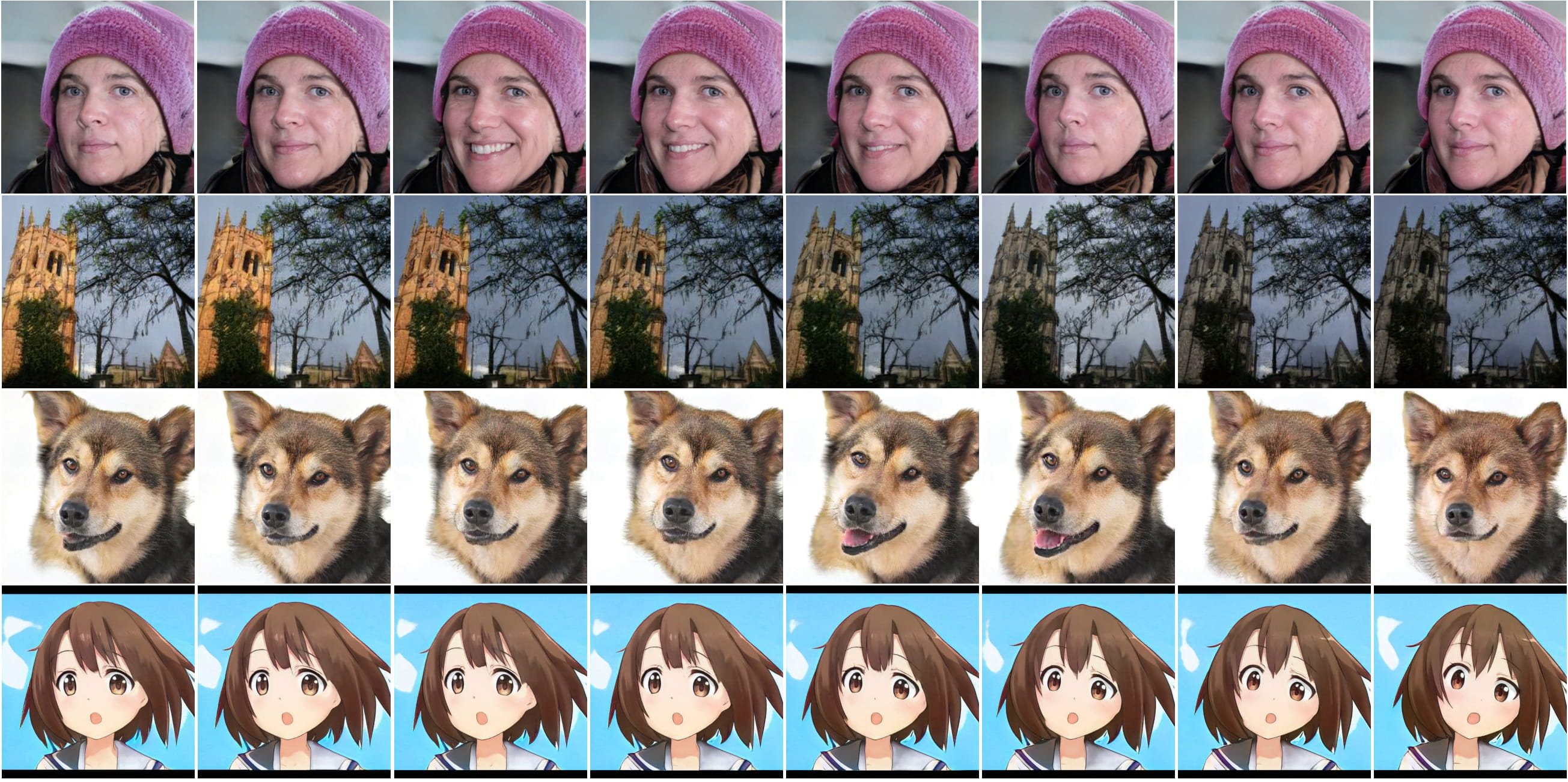}
\setlength{\tabcolsep}{0\linewidth}
\begin{tabular}{C{0.125\linewidth}C{0.125\linewidth}C{0.125\linewidth}C{0.125\linewidth}C{0.125\linewidth}C{0.125\linewidth}C{0.125\linewidth}C{0.125\linewidth}}
\small $t=2$&\small$t=4$&\small$t=6$&\small$t=8$&\small$t=10$&\small$t=12$&\small$t=14$&\small$t=16$
\end{tabular}
\caption{Example sequences for cross-domain video generation. {First Row}: (FFHQ, VoxCeleb). Second Row: (LSUN-Church, TLVDB). Third Row: (AFHQ-Dog, VoxCeleb). Fourth Row: (AnimeFaces, VoxCeleb). Images in the first and second rows have a resolution of $256\times256$, while the third and fourth rows have a resolution of $512\times512$.}\label{fig:cross-domain}
\end{center}
\end{figure}
\vspace*{-3mm}

\subsection{Cross-Domain Video Generation}\label{sec:cross_domin}
To demonstrate how our approach can disentangle motion from image content and transfer motion patterns from one domain to another, we perform several experiments on various datasets. More specifically, we use the StyleGAN2 model, pre-trained on the FFHQ~\citep{karras2019style}, AFHQ-Dog~\citep{choi2020starganv2}, AnimeFaces~\citep{branwen2019making}, and LSUN-Church~\citep{yu2015lsun} datasets, as the image generators. We learn human facial motion from VoxCeleb~\citep{nagrani2020voxceleb} and time-lapse transitions in outdoor scenes from TLVDB~\citep{shih2013data}. In these experiments, a pair such as (FFHQ, VoxCeleb) indicates that we synthesize videos with image content from FFHQ and motion patterns from VoxCeleb. 
% We generate videos with a resolution of $256\times256$, $1024\times1024$, $512\times512$, $512\times512$, and $256\times256$ for FFHQ, AFHQ-Dog, AnimeFaces, and LSUN-Church, respectively.
We generate videos with a resolution of $256\times256$ and $1024\times1024$ for FFHQ, $512\times512$ for AFHQ-Dog and AnimeFaces, and $256\times256$ for LSUN-Church.
Qualitative examples for (FFHQ, VoxCeleb), (LSUN-Church, TLVDB), (AFHQ-Dog, VoxCeleb), and (AnimeFaces, VoxCeleb) are shown in Fig.~\ref{fig:cross-domain}, depicting high-quality and temporally consistent videos (more videos, including results with BigGAN as the image generator, are shown in the Appendix).
% \textcolor{red}{Fig.~\ref{fig:cross-domain}}

\begin{figure}[h]
\centering
\begin{minipage}[t] {.49\linewidth}
\begin{center}
\includegraphics[width=1\linewidth]{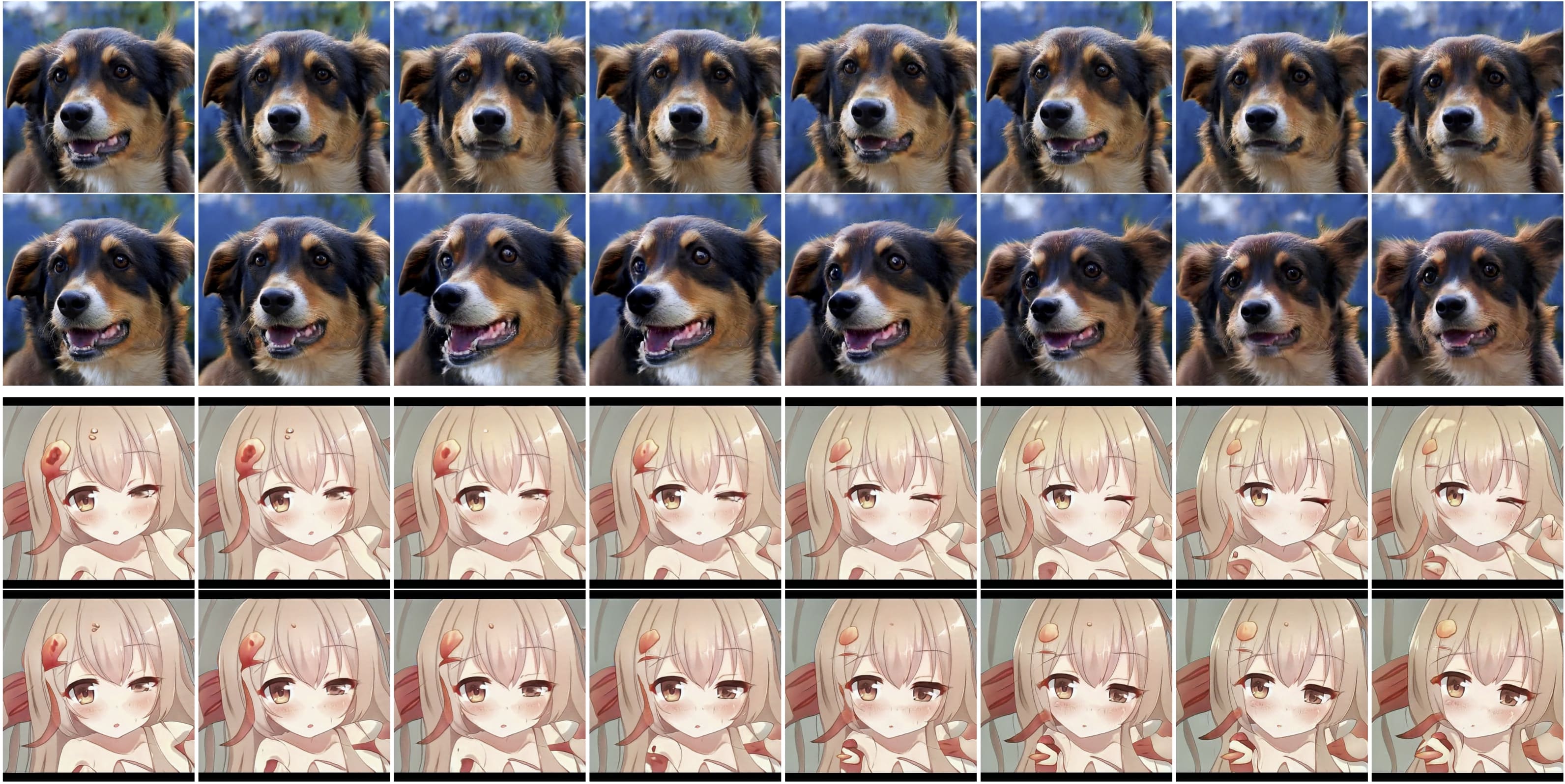}
\setlength{\tabcolsep}{0\linewidth}
\begin{tabular}{C{0.125\linewidth}C{0.125\linewidth}C{0.125\linewidth}C{0.125\linewidth}C{0.125\linewidth}C{0.125\linewidth}C{0.125\linewidth}C{0.125\linewidth}}
\tiny $t=2$& \tiny $t=4$&\tiny$t=6$&\tiny$t=8$&\tiny$t=10$&\tiny$t=12$&\tiny$t=14$&\tiny $t=16$
\end{tabular}
\caption{The first and second row (also the third and fourth row) share the same initial content code but with different motion codes.}\label{fig:sameC_diffM}
\end{center}
\end{minipage}\hfill
\begin{minipage}[t] {.49\linewidth}
\begin{center}
\includegraphics[width=1\linewidth]{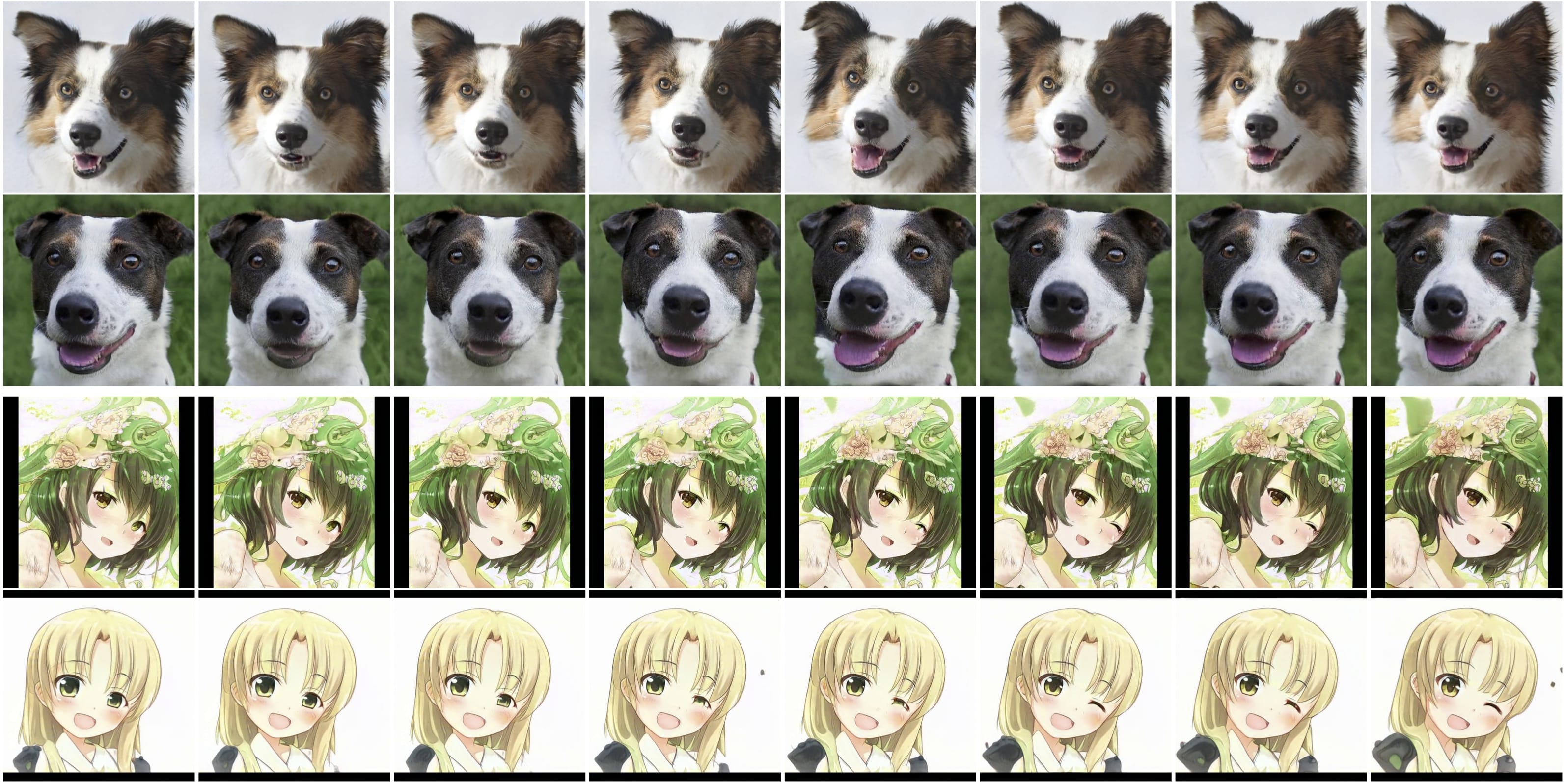}
\setlength{\tabcolsep}{0\linewidth}
\begin{tabular}{C{0.125\linewidth}C{0.125\linewidth}C{0.125\linewidth}C{0.125\linewidth}C{0.125\linewidth}C{0.125\linewidth}C{0.125\linewidth}C{0.125\linewidth}}
\tiny $t=2$& \tiny $t=4$&\tiny$t=6$&\tiny$t=8$&\tiny$t=10$&\tiny$t=12$&\tiny$t=14$&\tiny $t=16$
\end{tabular}
\caption{The first and second row (also the third and fourth row) share the same motion code but with different content codes.}\label{fig:sameM_diffC}
\end{center}
\end{minipage}
\end{figure}

% \textcolor{red}{To Do}
We also demonstrate how the motion and content are disentangled in Fig.~\ref{fig:sameC_diffM} and Fig.~\ref{fig:sameM_diffC}, which portray generated videos with the same identity but performing diverse motion patterns, and the same motion applied to different identities, respectively. We show results from (AFHQ-Dog, VoxCeleb) (first two rows) and (AnimeFaces, VoxCeleb) (last two rows) in these two figures.

% {\bf (FFHQ, VoxCeleb)}
% FFHQ is a high-quality image dataset of human faces, it consists of $70,000$ high-quality images at $1024\times1024$ resolution and contains considerable variation in terms of age, ethnicity and image background. VoxCeleb consists of short clips of human speech, extracted from interview videos uploaded to YouTube. There is a domain gap between the two datasets in terms of the image quality and xxx. We use StyleGAN2 generator pre-trained on FFHQ with resolution 256$\times$256.

% {\bf (AFHQ-Dog, VoxCeleb)}
% We further evaluate if our model can synthesize animal facial video with the knowledge from human face motions. AFHQ consists 15,000 high-quality images at 512$\times$512 resolution. We train a StyleGAN2 image generator on cat and dog images at resolution 256$\times$256, each domain has about 5000 training images.

% {\bf (AnimeFaces, VoxCeleb)}

% {\bf (LSUN-Church, TLVDB)}

% Diversity: same person different motion (ablation).
% Same motion applied to different content (FaceForensics, and FFHQ).

% \textbf{High-resolution video generation}.
% Datasets: FFHQ and AnimeFaces.

\subsection{Ablation Analysis}
We first report IS and FVD in Tab.~\ref{tab:ablation} for UCF-101 using the following methods:
\textit{w/o Eqn.~\ref{eqn:residual}} uses $\mathbf{z}_{t}=\mathbf{h}_{t}$ instead of estimating the residual as in Eqn.~\ref{eqn:residual}; 
\textit{w/o $D_\mathrm{I}$} omits the contrastive image discriminator $D_\mathrm{I}$ and uses the video discriminator $D_\mathrm{V}$ only for learning the motion generator; 
\textit{w/o $D_\mathrm{V}$} omits $D_\mathrm{V}$ during training; 
and \textit{Full-128} and \textit{Full-256} indicate that we generate videos using our full method with resolutions of $128\times128$ and $256\times256$, respectively. 
We resize frames for all methods to $128\times128$ when calculating IS and FVD.
% \textcolor{blue}{All videos, include the real one, are resized to $128\times 128$ before calculating IS and FVD.} 
The full method outperforms all others, proving the importance of each module for learning temporally consistent and high-quality videos.

% The proposed method has mixed multiple modules and losses for different purpose. To clarify the functions for modules and the role for losses, we conduct two kinds of ablation studies.

% {\bf Module Ablation.} In addition to the full model, we ablate out each single module and lead to three variants: 
% \begin{itemize}
%     \item (w/o PCA) Instead of estimating residual (Eqn.~\ref{eqn:residual}), we use $z_{t}=h_{t}$ directly.
%     \item (w/o 2D) We remove the 2D discriminator, let 3D discriminator to deal with content and motion only.
%     \item (w/o 3D) We remove the 3D discriminator to see how motion generator $G_{m}$ behaves under the guidance of sharp synthesizing and content matching.
% \end{itemize}

We perform further analysis of our cross-domain video generation on (FFHQ, VoxCeleb). We compare our full method (\textit{Full}) with two variants. \textit{w/o $\mathcal{L}_\mathrm{contr}$} denotes that we omit the contrastive loss (Eqn.~\ref{eqn:con}) from $D_\mathrm{I}$, and \textit{w/o $\mathcal{L}_\mathrm{m}$} indicates that we omit the mutual information loss (Eqn.~\ref{eqn:mut}) for the motion generator. The results in Tab.~\ref{tab:ablation-face} demonstrate that $\mathcal{L}_\mathrm{contr}$ is beneficial for learning videos with coherent content, as employing $\mathcal{L}_\mathrm{contr}$ results in lower ACD values and higher human preferences. $\mathcal{L}_\mathrm{m}$ also contributes to generating higher quality videos by mitigating motion synchronization. To validate the motion diversity, we show pairs of $9$ randomly generated videos from the two methods to users and ask them to choose which one has superior motion diversity, including rotations and facial expressions. User preference suggests that using $\mathcal{L}_\mathrm{m}$ increases motion diversity.

% {\bf Loss Ablation.} We show the importance of contrastive loss and mutual information loss with two variants:
% \begin{itemize}
%     \item (w/o contrastive) We remove the contrastive loss $L_{con}$ and feature matching loss $L_{f}$, as a result, the 2D discriminator only take care of frame quality.
%     \item (w/o mutual) We disable the mutual information loss $L_{m}$, motion generator $G_{m}$ can decide how to use noise vector $e_{t}$ in Eqn.~\ref{eqn:lstm}
% \end{itemize}
% \textcolor{blue}{We evaluate the ACD for each method, as shown in Table \ref{tab:ablation-face}, contrastive loss leads to more consistent content and better ACD score. The human preference also agrees that the content consistency is related to visual pleasant. We also compare the full model with w/o mutual, for each method, we sample 9 videos with the same first frame, and ask the Amazon AMT worker to choose which 9-grid video has better diversity. The results show that mutual information loss can improve the diversity significantly, while not affect the content consistency.} 

% Quantitative experiments on UCF-101 and qualitative on FFHQ.

\begin{table}[h]
\begin{minipage} {.45\linewidth}
\centering
    \caption{\label{tab:ablation} Ablation study on UCF-101.}
    \begin{tabular}{c|cc}
    \toprule 
    Method & IS ($\uparrow$)  & FVD ($\downarrow$)  \\ \midrule
    w/o Eqn.~\ref{eqn:residual}  & 28.20 %28.1963107581873
    & 790.87 
    %790.86804
    \\
    w/o $D_\mathrm{I}$ & 33.22 & 796.67 \\
    w/o $D_\mathrm{V}$ & 33.84 & 867.43 \\
    Full-128 & 32.36
    %32.36184952499383
    & 838.09 
    %838.0933 
    \\
    Full-256 & {\bf 33.95}  & {\bf 700.00} \\
    \bottomrule
    \end{tabular} 
\end{minipage}\hfill
\begin{minipage}{.5\linewidth}
\centering
    \caption{\label{tab:ablation-face} Ablation study on (FFHQ, VoxCeleb).}
    %\centering
    \begin{tabular}{c|ccc}
    %\centering
    \toprule
    Method & w/o $\mathcal{L}_\mathrm{contr}$ & w/o $\mathcal{L}_\mathrm{m}$ & Full \\ \midrule
    ACD ($\downarrow$) & 0.5328 
    %0.5327962494495144
    & 0.5158 
    %0.5158061276488145 
    & \textbf{0.4353} 
    %0.4353259978224893 
    \\ \bottomrule
    \end{tabular}
    \begin{tabular}{c|c}
    \toprule
    Method & Human Preference (\%) \\ \midrule
   Full \textit{vs} w/o $\mathcal{L}_\mathrm{contr}$ & \textbf{68.3} / 31.7  \\
 Full \textit{vs} w/o $\mathcal{L}_\mathrm{m}$ & \textbf{64.4} / 35.6      \\    \bottomrule
    \end{tabular}
\end{minipage}
\end{table}

\subsection{{Long Sequence Generation}}
{Due to the limitation of computational resources, we train MoCoGAN-HD to synthesize $16$ consecutive frames. However, we can generate longer video sequences during inference by applying the following two ways.}

\begin{figure}[h]
 \begin{center}
    \includegraphics[width=1\linewidth]{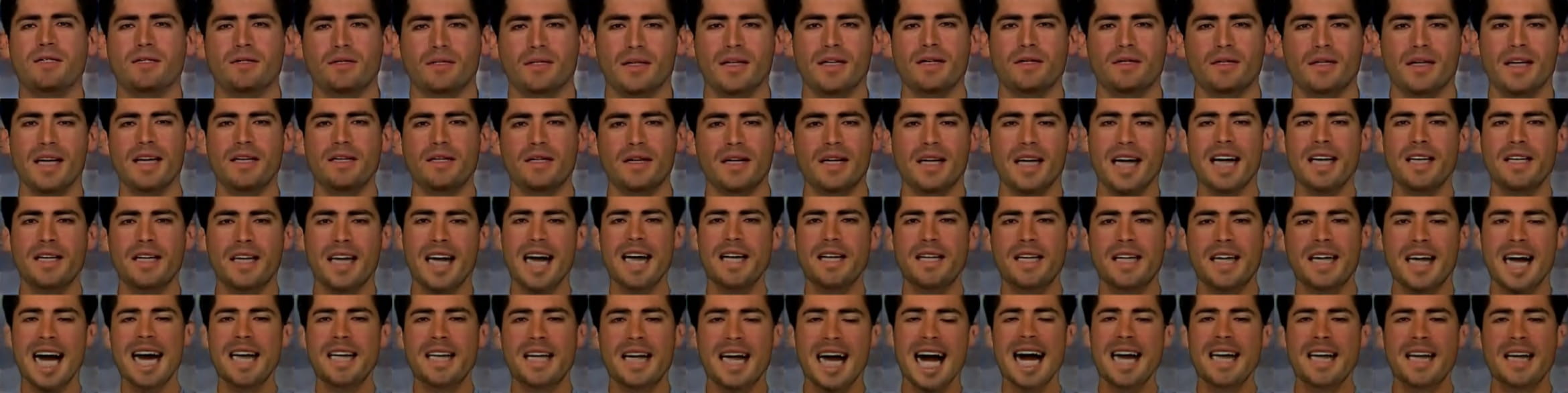}
  \end{center}
  \caption{The generation of a $64$-frame video using a model trained with $16$-frame on FaceForensics.}\label{fig:face64}
% \end{figure}
\end{figure}

{\bf Motion Generator Unrolling.}
For motion generator, we can run the LSTM decoder for more steps to synthesize long video sequences. In Fig.~\ref{fig:face64}, we show a synthesized video example of $64$ frames using the model trained on the FaceForensics dataset. Our method is capable to synthesize videos with more frames than the number of frames used for training. 
% We noticed TGANv2~\citep{TGAN2020} sometimes generated broken videos longer than 16 frames which is in contrast to us. We hypothesis it is due to the residual design in the LSTM decoder, which makes the prediction easier. See Appendix for more results.

\begin{figure}[h]
 \begin{center}
    \includegraphics[width=1\linewidth]{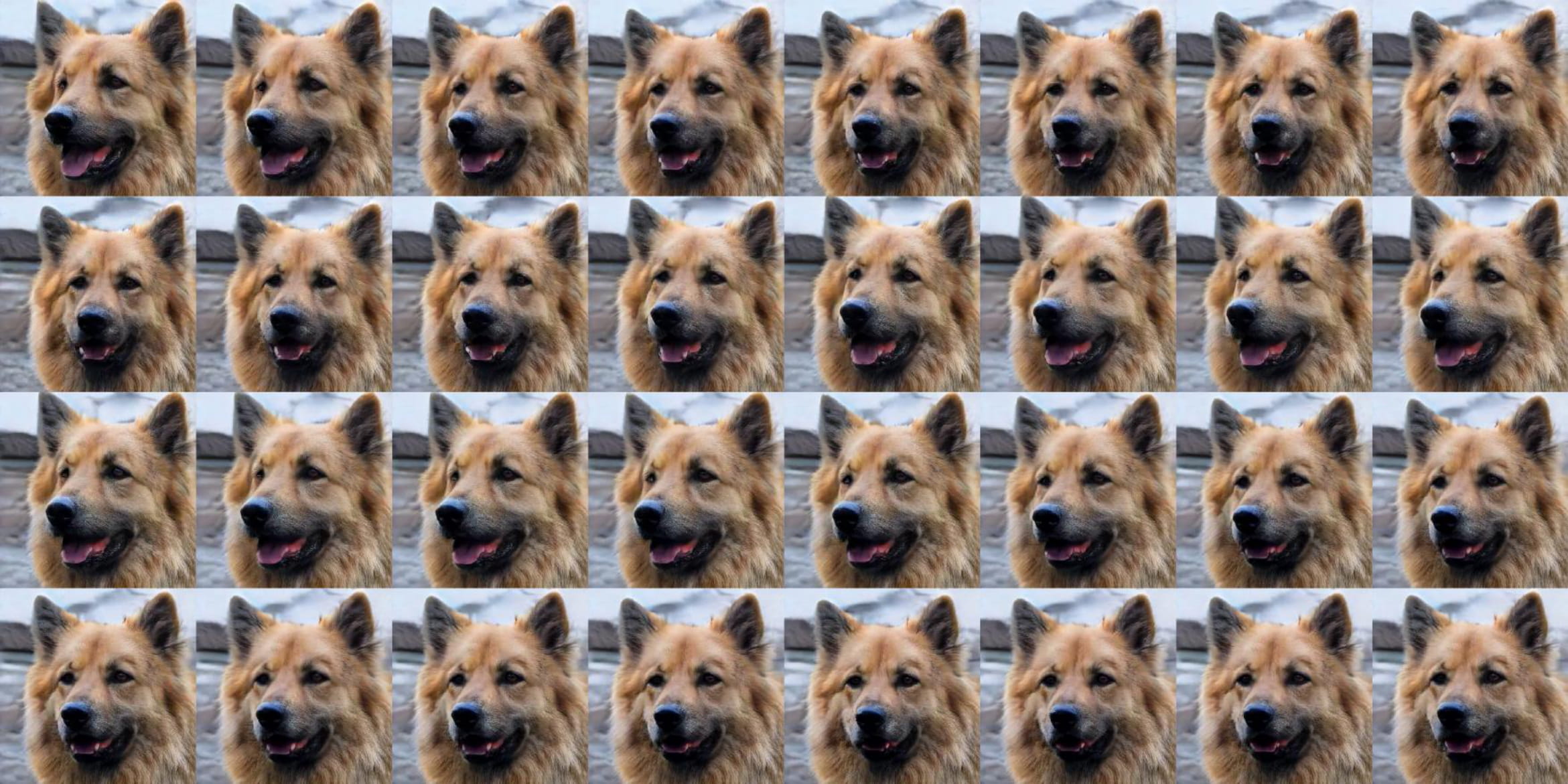}
  \end{center}
  \caption{The generation of a $32$-frame video on (AFHQ-Dog, VoxCeleb) by doing the interpolation on motion trajectory.}\label{fig:dog32}
% \end{figure}
\end{figure}

{\bf Motion Interpolation.}
We can do interpolation on the motion trajectory directly to synthesize long videos. Fig.~\ref{fig:dog32} shows an interpolation example of $32$-frame on (AFHQ-Dog, VoxCeleb) dataset. 
%Compared to 16-frame sample (odd frames), the 32-frame interpolation has more smooth motion.

\section{Conclusion}
In this work, we present a novel approach to video synthesis. Building on contemporary advances in image synthesis, we show that a good image generator and our framework are essential ingredients to boost video synthesis fidelity and resolution. The key is to find a meaningful trajectory in the image generator's latent space. This is achieved using the proposed motion generator, which produces a sequence of motion residuals, with the contrastive image discriminator and video discriminator. This disentangled representation further extends applications of video synthesis to content and motion manipulation and cross-domain video synthesis. The framework achieves superior results on a variety of benchmarks and reaches resolutions unattainable by prior state-of-the-art techniques.

% \clearpage
\bibliography{iclr2021_conference}
\bibliographystyle{iclr2021_conference}

\appendix

\newpage
\section{Additional Details for the Framework}
\subsection{Additional Details for the Motion Generator}
To use StyleGAN2~\citep{karras2020analyzing} as the image generator, 
we randomly sample $1,000,000$ latent codes from the input space $\mathcal{Z}$ and send them to the 8-layer MLPs to get the latent codes in the space of $\mathcal{W}$. 
Each latent code is a $512$-dimension vector. 
We perform PCA on these $1,000,000$ latent codes and select the top $384$ principal components to form the matrix $\mathbf{V}\in \mathbb{R}^{384\times512}$, which is used to model the motion residuals in Eqn.~\ref{eqn:residual}.
The LSTM encoder and the LSTM decoder in the motion generator both have an input size of $512$ and a hidden size of $384$. The noise vector $\epsilon_{t}$ in Eqn.~\ref{eqn:lstm} is also a $512$-dimension vector, and the network $H$ in Eqn.~\ref{eqn:mut} is a 2-layer MLPs with $512$ hidden units in each of the two fully-connected layers. 

For BigGAN~\citep{brock2018large}, we sample the latent code directly from the space of $\mathcal{Z}$. 
% Linear(384,512)$\rightarrow$ReLU()$\rightarrow$Linear(512,512).

% \textcolor{blue}{
% For BigGAN, we let the latent code $z$ works in $\mathcal{Z}$ space.  
% We know $z$ is sampled from the normal distribution, so instead of using PCA, we let hidden state modeling motion residual directly. Eqn.~\ref{eqn:residual} can be modified as:
% \begin{equation}\label{eqn:residual-biggan}
%     \mathbf{z}_{t} = \mathbf{z}_{t-1} + \lambda\cdot  \mathbf{h}_{t}, ~~~ t = 2, 3, \cdots, n,
% \end{equation}
% in this case, $\lambda=1$. The LSTM encoder and decoder have input size and hidden size both of $120$. The noise vector $\epsilon_{t}$ and $H$ in Eqn.~\ref{eqn:mut} also has size $120$.}

% \textcolor{blue}{
% We train BigGAN generator on FFHQ dataset with the public PyTorch code\footnote{https://github.com/ajbrock/BigGAN-PyTorch}. We train $128\times 128$ resolution model for XXX iterations and select the last checkpoint as the FID calculated by PyTorch is not accurate.
% }

\subsection{Additional Details for the Discriminators}
\subsubsection{Video Discriminator}
The input images for the video discriminator $D_\mathrm{V}$ are processed at two scales. We downsample the output images from the image generator to the resolution of $128\times128$ and $64\times64$. {For in-domain video synthesis,} the input sequences for $D_\mathrm{V}$ have the shape of $6\times(n-1)\times128\times128$ and $6\times(n-1)\times64\times64$, where $n$ is the sequence length used for training. For each of the $(n-1)$ subsequent frames, we concatenate the RGB channels of both the first frame and that subsequent frame, resulting in a $6$-channel input. {For cross-domain video synthesis, the input sequences for $D_\mathrm{V}$ have the shape of $3\times n\times128\times128$ and $3\times n\times64\times64$, as the concatenation of the first frame will make the discriminator aware the domain gaps.} Details for $D_\mathrm{V}$ are shown in Tab.~\ref{tab:3d-dis-param}.
% The adversarial training loss is updated by by relativistic average loss~\citep{jolicoeur2018relativistic}.

% \textcolor{blue}{The hyperparameters of the 3D video discriminator $D_{v}$ are shown in Table~\ref{tab:3d-dis-param}. We implement $D_{v}$ as a 2-parts multi-scale discriminator. After downsample generator outputs into $128$ resolution video $v$, We send $v$ to $D_{v}$ part 1 to get the output $d_{out_1}$. Then, we downsample $v$ along height and width to get video $v_{down}$ and send $v_{down}$ to $D_{v}$ part 2 to get $d_{out_2}$. We use $d_{out_1}$ and $d_{out_2}$ to update $D_{v}$ by relativistic average loss (RaGAN loss)~\citep{jolicoeur2018relativistic}.}

% \textcolor{blue}{Note that 1) $v$ has size 6$\times$15$\times$128$\times$128 as we attach the first frame of a 3$\times$16$\times$128$\times$128 video to every other frames. 2) $D_{v}$ is fully convolutional, so it can take input with any size, we expect an image size larger than 128 may benefit the training (We also downsample $1,024$ resolution video to $128$ before sending it to $D_{v}$).}
\begin{table}[ht]
\centering
% \begin{minipage}[t] {.45\linewidth}
    \caption{\label{tab:3d-dis-param} The network architecture for video discriminator.}
    \centering
    % \resizebox{\textwidth}{!}{
    \begin{tabular}{cccccc}
    \toprule
    Operation & Kernel & Strides  & \# Channels & Norm Type & Nonlinearity   \\
    \midrule
    % $D_{v}(v)$ part1~-~6$\times$15$\times$128$\times$128 input & & & & \\
    Conv3d & 4$\times$4 & 2 & 64 & - & Leaky ReLU (0.2)\\
    Conv3d & 4$\times$4 & 2 & 128 & InstanceNorm3d & Leaky ReLU (0.2)\\
    Conv3d & 4$\times$4 & 2 & 256 & InstanceNorm3d & Leaky ReLU (0.2)\\
    Conv3d & 4$\times$4 & 1 & 512 & InstanceNorm3d & Leaky ReLU (0.2)\\
    Conv3d & 4$\times$4 & 1 & 1 & - & -\\
    \bottomrule
    % $D_{v}(v_{down})$ part2~-~6$\times$15$\times$64$\times$64 input & & & & \\
    % Conv3d & 4$\times$4 & 2 & 64 & - & Leaky ReLU (0.2)\\
    % Conv3d & 4$\times$4 & 2 & 128 & InstanceNorm3d & Leaky ReLU (0.2)\\
    % Conv3d & 4$\times$4 & 2 & 256 & InstanceNorm3d & Leaky ReLU (0.2)\\
    % Conv3d & 4$\times$4 & 1 & 512 & InstanceNorm3d & Leaky ReLU (0.2)\\
    % Conv3d & 4$\times$4 & 1 & 1 & - & -\\
    % \bottomrule
    \end{tabular}
    % }
\end{table}

\subsubsection{Image Discriminator}
The image discriminator $D_\mathrm{I}$ has an architecture based on that of the BigGAN discriminator, except that we remove the self-attention layer. The feature extractor $F$ used for contrastive learning has the same architecture as $D_\mathrm{I}$, except that it does not include the last layer of $D_\mathrm{I}$ but has two additional fully connected (FC) layers as the projection head. The number of hidden units for these two FC layers are both $256$.

Here we describe in more detail the image augmentation and memory bank techniques used for conducting contrastive learning. 

% We train image generator with resolution ranges from $128\times128$ to $1024\times1024$. To save memory, we resize all generated frames into $128\times128$ with 3D average pooling. As the image generator is pre-trained and fixed during video generation training, this design will not affect frame quality to much. For image discriminator, we adopt a no-attention BigGAN discriminator, which is identical to the BigGAN $128\times128$ expect the self-attention layer. To facilitate contrastive learning, following~\citep{chen2020big}, we use a 2-layer MLP as the projection head.
% \textcolor{red}{architecture details in a form or figure}
% For image discriminator, we adopt hinge loss as in BigGAN.
% Here we introduce more details for the implementation of contrastive loss.

{\bf Image Augmentation.}
We perform data augmentation on images to create positive and negative pairs.
We normalize the images to $\left [ -1, 1 \right ]$ and apply the following augmentation techniques.
% by considering the following five kinds of approaches:
\begin{itemize}
\item {\bf Affine.} We augment each image with an affine transformation defined with three random parameters: rotation $\alpha_{r}\in\mathcal{U}(-180,180)$, translation $\alpha_{t}\in\mathcal{U}(-0.1,0.1)$, and scale $\alpha_{s}\in\mathcal{U}(0.95,1.05)$.
\item {\bf Brightness.} We add a random value $\alpha_b\sim\mathcal{U}(-0.5,0.5)$ to all channels of each image.
\item {\bf Color.} We add a random value $\alpha_c\sim\mathcal{U}(-0.5,0.5)$ to one randomly-selected channel of each image.
\item {\bf Cutout}~\citep{devries2017improved}{\bf.} We mask out pixels in a random subregion of each image to $0$. Each subregion starts at a random point and with size $(\alpha_mH, \alpha_mW)$, where $\alpha_m\sim\mathcal{U}(0,0.25)$ and $(H, W)$ is the image resolution.
\item {\bf Flipping}. We horizontally flip the image with the probability of $0.5$.
\end{itemize}

{\bf Memory Bank.}
It has been shown that contrastive learning benefits from large batch-sizes and negative pairs~\citep{chen2020big}. To increase the number of negative pairs, we incorporate the memory mechanism from MoCo~\citep{he2020momentum}, which designates a memory bank to store negative examples. More specifically, we keep an exponential moving average of the image discriminator, and its output of \emph{fake} video frames are buffered as negative examples. We use a memory bank with a dictionary size of $4,096$.

\section{More Details for Experiments}\label{app:experiments}

% Training machine: 8 Tesla V100 for resolution as $256\times256$ and $512\times512$.
% \textcolor{red}{machine type} for resolution as $1024\times1024$

% \textcolor{red}{Not sure whether necessary: an ablation study on $\mathcal{Z}$ space (normal distribution) for StyleGAN2.}

% {\bf Baseline details.}

{\bf Image Generators.}
We train the unconditional StyleGAN2 models from scratch on the UCF-101, FaceForensics, Sky Time-lapse, and AFHQ-Dog datasets. 
% For the video datasets, we train with their frame images. 
We train the image generators with the official Tensorflow code\footnote{https://github.com/NVlabs/stylegan2} and select the checkpoints that obtain the best Fr$\mathrm{\acute{e}}$chet inception distance (FID)~\citep{heusel2017gans} score to be used as the image generators.
% \textcolor{blue}{The Frechet inception distance (FID)~\citep{heusel2017gans} is the measure for the quality of the generated examples that uses 2nd order information of the final layer of the inception model applied to the examples. It evaluates the 2-Wasserstein distance between two distribution p1 and p2 assuming they are both multivariate Gaussian distributions.}
% as it has shown that FID is better than ICP in evaluating image quality and diversity~\citep{heusel2017gans}. 
The FID score of each image generator is shown in Table~\ref{tab:g-fid}. 
% The Tensorlow models are transfered to PyTorch using online code\footnote{https://github.com/rosinality/stylegan2-pytorch}.
% We train the image generators with official code\footnote{https://github.com/NVlabs/stylegan2} and transfer the checkpoint to PyTorch using online codes\footnote{https://github.com/rosinality/stylegan2-pytorch}. 
For FFHQ, AnimeFaces, and 
LSUN-Church, we simply use the released pre-trained models.
% on FFHQ (256, 1024), LSUN-church and AnimeFaces for cross domain video generation.

We also train an unconditional BigGAN model on the FFHQ dataset using the public PyTorch code\footnote{https://github.com/ajbrock/BigGAN-PyTorch}. We train a model with resolution $128\times 128$ and select the last checkpoint as the image generator.
% as the FID calculated by PyTorch is not accurate.

\begin{table}[h]
\centering
    \caption{\label{tab:g-fid} FID of our trained StyleGAN2 models on different datasets.}
    \begin{tabular}{c|cccc}
    \toprule
     & UCF-101 & FaceForensics & Sky Time-lapse & AFHQ-Dog \\ 
    \midrule
    FID  & 45.63 & 10.99 & 10.80 & 7.85 \\
    % kimg & 11854 & 8765 & 12633 & 5655 \\
    \bottomrule
    \end{tabular} 
\end{table}

{\bf Training Time.} 
% As shown in Table~\ref{tab:g-fid}, all models pre-trained by us use less than $13,000k$ images, which can be finished $<2$ days on 8 Tesla V100. Depends on the dataset, the video generation model needs $1.5\sim 3$ days to train, which results in $3\sim 5$ days in total.
We train each image generator for UCF-101, FaceForensics, Sky Time-lapse, and AFHQ-Dog in less than 2 days using 8 Tesla V100 GPUs. For FFHQ, AnimeFaces, and LSUN-Church, we use the released models with no training cost.
The training time for video generators ranges from $1.5\sim 3$ days depending on the datasets {(Due to the memory issue, the training for generating videos with resolution of $1,024\times1,024$ was done on 8 Quadro RTX 8000, with 5 days)}. The total training time for all the datasets is $1.5\sim 5$ days and the estimated cost for training on Google Cloud is \$0.7K$\sim$\$2.3K.

{\bf Implementation Details.} We implement our experiments with PyTorch 1.3.1 and also tested them with PyTorch 1.6. We use the Adam optimizer~\citep{kingma2014adam} with a learning rate of $0.0001$ for $G_\mathrm{M}$, $D_\mathrm{V}$, and $D_\mathrm{I}$ in all experiments. 
% The image generator $G_{I}$ is pre-trained and fixed in the video generation training. 
In Eqn.~\ref{eqn:residual}, we set $\lambda=0.5$ for conventional video generation tasks and use a smaller $\lambda=0.2$ for cross-domain video generation, as it improves the content consistency.
In Eqn.~\ref{eqn:full}, we set $\lambda_\mathrm{m}=\lambda_\mathrm{contr}=\lambda_\mathrm{f}=1$. Grid searching on these hyper-parameters could potentially lead to a performance boost.
For TGANv2, we use the released code\footnote{https://github.com/pfnet-research/tgan2} to train the models on UCF-101 and FaceForensics using 8 Tesla V100 with $16$GB of GPU memory.
%\textcolor{red}{Not sure whether we need this in the main article as it can be confusing}\textcolor{blue}{Note that previous work shows the agreement between human raters is close to random when the FVD scores for two methods are smaller than 50~\citep{unterthiner2018towards}.  Considering the backbone model of FVD is trained on Kinetics dataset~\citep{carreira2017quo}, it is not sensitive to facial related attributes such as identity consistency}
 
{\bf Video Prediction.}
For video prediction, we predict consecutive frames, given the first frame $\mathbf{x}$ from a test video clip as the input.
% Our model don't have an encoder to find the inverse code in latent space, instead, we use the code\footnote{https://github.com/rosinality/stylegan2-pytorch}
We find the inverse latent code $\hat{\mathbf{z}}_1$ for $\mathbf{x}_1$ by minimizing the following objective:
\begin{equation}\label{eqn:projector}
    \hat{\mathbf{{z}}}_1=\underset{\hat{\mathbf{z}}_1}{\arg\min}\left \| \mathbf{x}_1-G_\mathrm{I}(\hat{\mathbf{z}}_1) \right \|_{2} +\lambda_\mathrm{vgg} \left \| F_\mathrm{vgg}(\mathbf{x}_1)-F_\mathrm{vgg}(G_\mathrm{I}(\hat{\mathbf{z}}_1)) \right \|_{2},
\end{equation}
where $\lambda_\mathrm{vgg}$ is the weight for perceptual loss~\citep{johnson2016perceptual}, $F_\mathrm{vgg}$ is the VGG feature extraction model~\citep{simonyan2014very}. We set $\lambda_\mathrm{vgg}=1$ and optimize Eqn.~\ref{eqn:projector} for $20,000$ iterations.
We take $\hat{\mathbf{z}}_1$ as the input to our model for video prediction.
 
{\bf AMT Experiments.}
% \textcolor{red}{How do we select workers.}
We present more details on the AMT experiments for different experimental settings and datasets. For each experiment, we run $5$ iterations to get the averaged score.
\begin{itemize}
    \item \emph{FaceForensics, Ours vs TGANv2.} We randomly select $300$ videos from each method and ask users to select the better one from a pair of videos. 
    \item 
\emph{Sky Time-lapse, Ours vs DTVNet.} 
We compare our method with DTVNet on the video prediction task.
The testing set of Sky Time-lapse dataset includes $2,815$ short video clips. Considering that many of these video clips share similar content and are sampled from $148$ long videos, we select $148$ short videos with different content for testing. For these videos, we perform inversion (Eqn.~\ref{eqn:projector}) on the first frame to get the latent code and generate videos. For DTVNet, we use the first frame directly as input to produce their results. We ask users to chose the one with better video quality from a pair of videos generated by our method and DTVNet. The results shown in Tab.~\ref{tab:dtvnet_human} demonstrate the clear advantage of our approach. 

\begin{table}[h]
\centering
\caption{Human evaluation experiments on Sky Time-lapse dataset.}\label{tab:dtvnet_human}
\begin{tabular}{c|c}
    \toprule 
    Method & Human Preference (\%)  \\ \midrule
     Ours / DTVNet   &  \textbf{77.3} / 22.7 \\
    \bottomrule
\end{tabular}
\end{table}

\item \emph{FFHQ, Full vs w/o $\mathcal{L}_\mathrm{contr}$.} We randomly sample $200$ videos generated by each method and ask users to select the more realistic one from a pair of videos. 

\item \emph{FFHQ, Full vs w/o $\mathcal{L}_\mathrm{m}$.}
For each method, we use the same content code $\mathbf{z}_1$ to generate $9$ videos with different motion trajectories, and organize them into a $3\times3$ grid. To conduct AMT experiments, we randomly generate $50$ $3\times3$ videos for each method and ask users to choose the one with higher motion diversity from a pair of videos. 
% The method with more motion diversity should include more rotation and facial expression.  

\end{itemize}

% \textcolor{red}{26 workers, ours $1,092$, theirs $392$.}

% We use the videos generated by the video prediction task on test set. The Sky Time-lapse test dataset has $2,815$ video clips, most of them have similar content with other clips. For example, $\mathrm{ZfUcldPsLVo\_3\_21.mp4}$ and $\mathrm{ZfUcldPsLVo\_3\_22.mp4}$ are two clips from same video. We parse the video name by $token = v\_name.split('\_')$, and randomly select one video clip from all clips where share the same first two tokens, which results in $148$ video clips. For those clips, we use the same video prediction method mentioned in the main paper to produce our results. While we let DTVNet take real first frame as input to produce their results. We ask users to select the better one from a pair of videos. 
% \textcolor{red}{32 workers, ours $569$, theirs $167$.}

% \textcolor{red}{25 workers, ours $683$, theirs $317$.}

% We use the same latent code as input for $9$ times to generate $9$ video clips with same first frame but different later motion. Then combine 9 video clips into a 3$\times$3 grid video. We randomly generate $50$ 3$\times$3 grid videos for each methods to ask users to select the one video with more motion diversity (more rotations, more facial expressions) from a pair of videos and perform the experiments for $5$ times to get the average score. 
% \textcolor{red}{14 workers, ours $161$, theirs $89$.}

{\bf Cross-Domain Video Generation.}
{We provide more details on the image and video datasets.}

\begin{itemize}
\item Image Datasets:
\begin{itemize}
\item \emph{FFHQ}~\citep{karras2019style} consists of $70,000$ high-quality face images at $1024\times1024$ resolution with considerable variation in terms of age, ethnicity, and background.
\item \emph{AFHQ-Dog}~\citep{choi2020starganv2} contains $5,239$ high-quality dog images at $512\times512$ resolution with both training and testing sets.
\item \emph{AnimeFaces}~\citep{branwen2019making} includes $ 2,232,462$ anime face images at $512\times512$ resolution.
\item \emph{LSUN-Church}~\citep{yu2015lsun} includes $126,227$ in-the-wild church images at $256\times256$ resolution.
\end{itemize}
\item Video Datasets:
\begin{itemize}
    \item \emph{VoxCeleb}~\citep{nagrani2020voxceleb} consists of $22,496$ short clips of human speech, extracted from interview videos uploaded to YouTube.
    \item \emph{TLVDB}~\citep{shih2013data} includes $463$ time-lapse videos, covering a wide range of landscapes and cityscapes.
\end{itemize}
\end{itemize}

For the video datasets, we randomly select $32$ consecutive frames from training videos and select every other frame to form a 16-frame sequence for training.

% \textcolor{blue}{We generate our training video clip by following steps: randomly select 32 consecutive frames from training video, then select 1 frame from every two frames to form a 16-frame clip for training. In Fig.~XXX we show we can interpolate 16 frames to 32 frames on AFHQ-DOG dataset.}

\section{More Video Results}\label{app:more_results}
In this section, we provide more qualitative video results generated by our approach.
We show the thumbnail from each video in the figures. Full resolution videos are in the supplementary material. We also provide an HTML page to visualize these videos.
% Each figure can be played by clicking on it when viewed via Acrobat Reader. Note the figures contain downsampled images, and the full resolution videos are in the supplementary material.

\begin{figure}[h]
\begin{minipage}[t]{.49\linewidth}
\centering
% \makebox[0mm][s]{
% \includegraphics[width=0.8\linewidth]{gif/ucf/1.jpg}
% }
% \animategraphics[width=0.8\linewidth]{10}{gif/ucf/}{1}{16}
\includegraphics[width=0.8\linewidth]{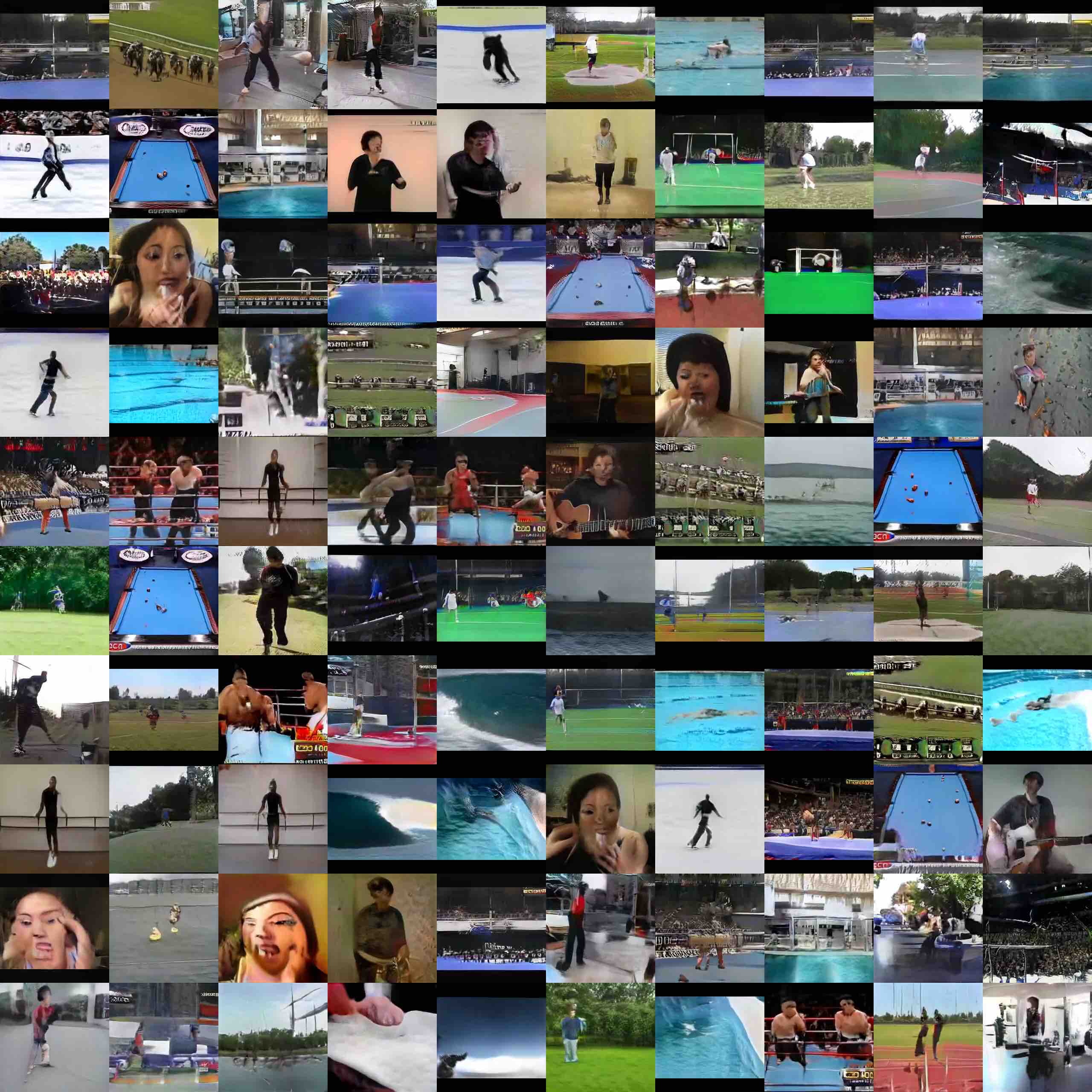}
\caption{Example videos generated by our approach on the UCF-101 dataset.}\label{fig:ucf_qualitative}
\end{minipage}\hfill
\begin{minipage}[t]{.49\textwidth}
\centering
% \makebox[0mm][s]{
% \includegraphics[width=0.8\linewidth]{gif/faceforensics/1.jpg}}
% \animategraphics[width=0.8\linewidth]{10}{gif/faceforensics/}{1}{16}
\includegraphics[width=0.8\linewidth]{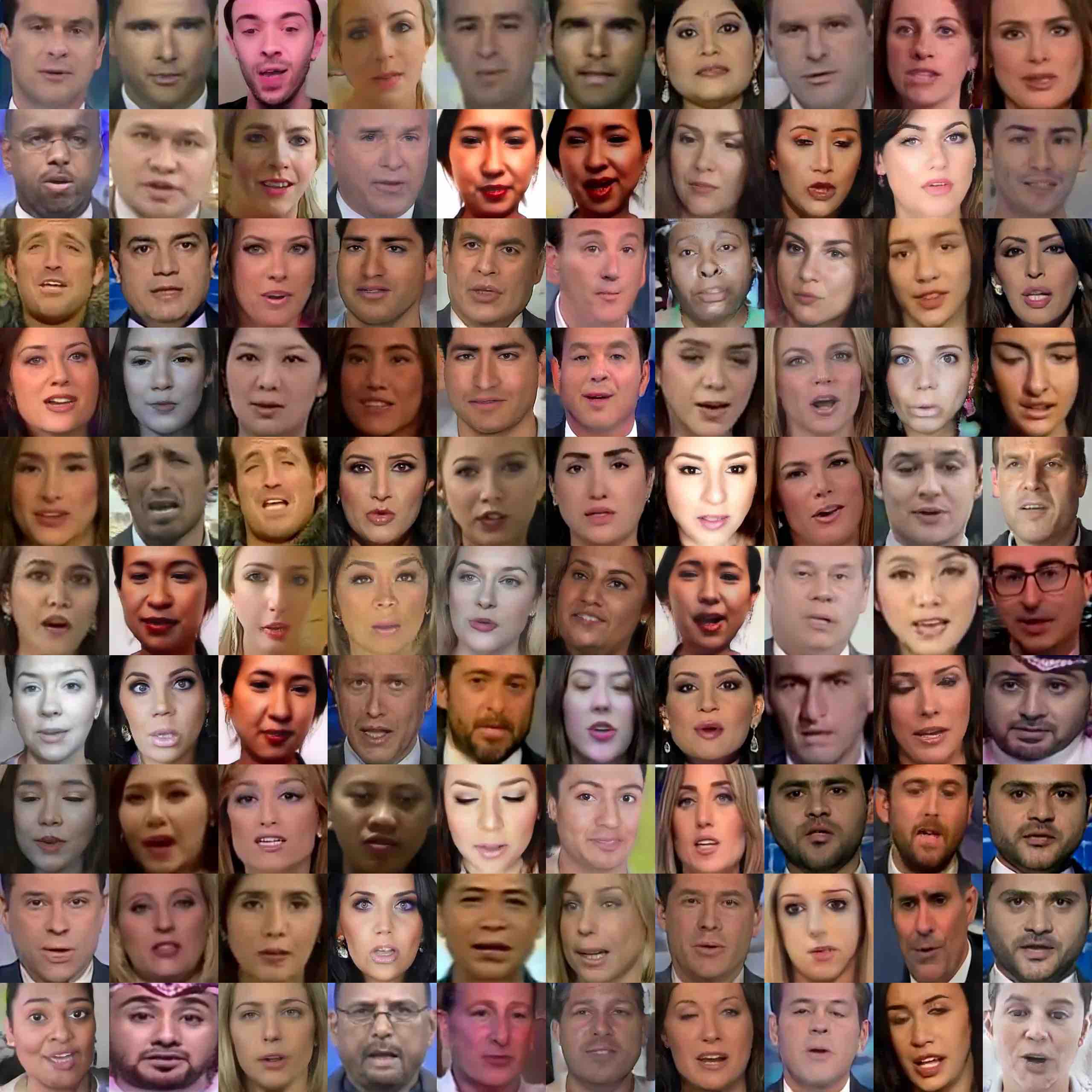}
\caption{Example videos generated by our approach on the FaceForensics dataset.}\label{fig:face_appendix}
\end{minipage}
\end{figure}

\textbf{UCF-101.} In Fig.~\ref{fig:ucf_qualitative}, we show videos generated by our approach on the UCF-101 dataset.

\begin{figure}[h]
\begin{minipage}[t]{.49\linewidth}
\centering
% \makebox[0mm][s]{
% \includegraphics[width=0.8\linewidth]{gif/faceforensics32/1.jpg}
% }
% \animategraphics[width=0.8\linewidth]{10}{gif/faceforensics32/}{1}{32}
\includegraphics[width=0.8\linewidth]{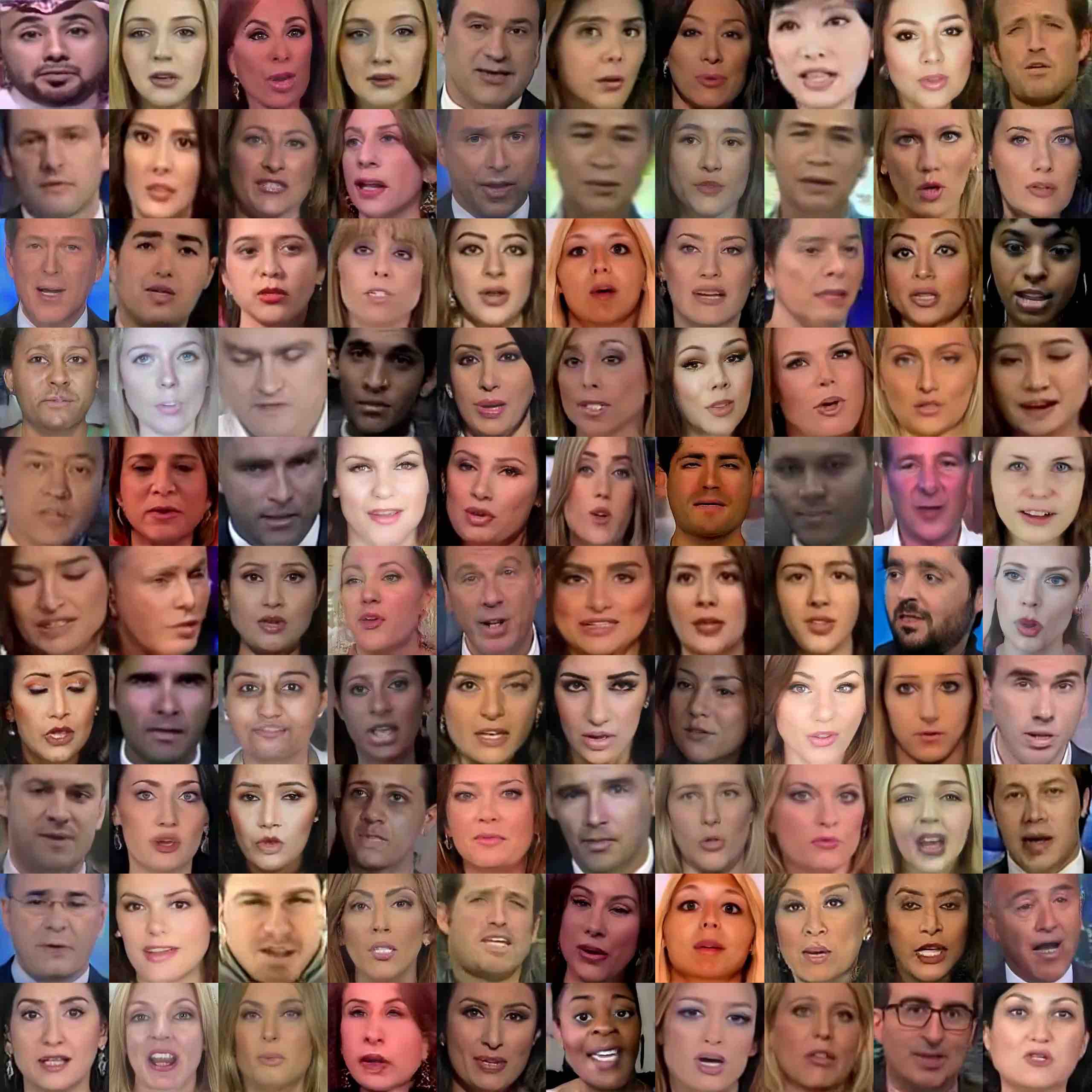}
\caption{The generated videos on the FaceForensics dataset consisting of 32 frames.}\label{fig:face_32_appendix}
\end{minipage}\hfill 
\begin{minipage}[t]{.49\textwidth}
\centering
% \makebox[0mm][s]{
% \includegraphics[width=0.8\linewidth]{gif/faceforensics64/1.jpg}}
% \animategraphics[width=0.8\linewidth]{10}{gif/faceforensics64/}{1}{64}
\includegraphics[width=0.8\linewidth]{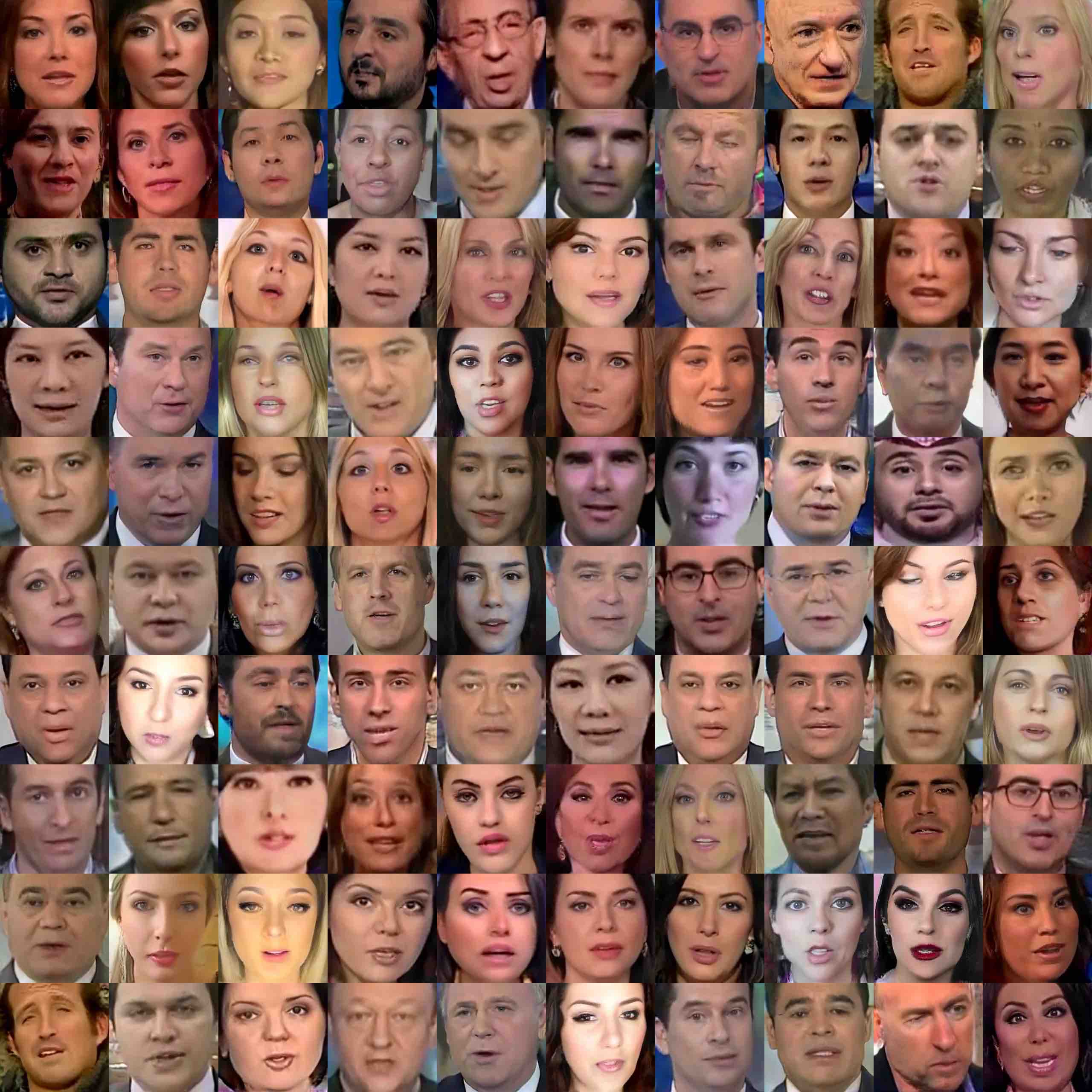}
\caption{The generated videos on the FaceForensics dataset consisting of 64 frames.}\label{fig:face_64_appendix}
\end{minipage}
\end{figure}

\begin{figure}[h]
\begin{minipage}[t]{.49\linewidth}
\centering
% \makebox[0mm][s]{
% \includegraphics[width=0.8\linewidth]{gif/faceforensics_samecont_diffmotion/1.jpg}}
% \animategraphics[width=0.8\linewidth]{10}{gif/faceforensics_samecont_diffmotion/}{1}{16}
\includegraphics[width=0.8\linewidth]{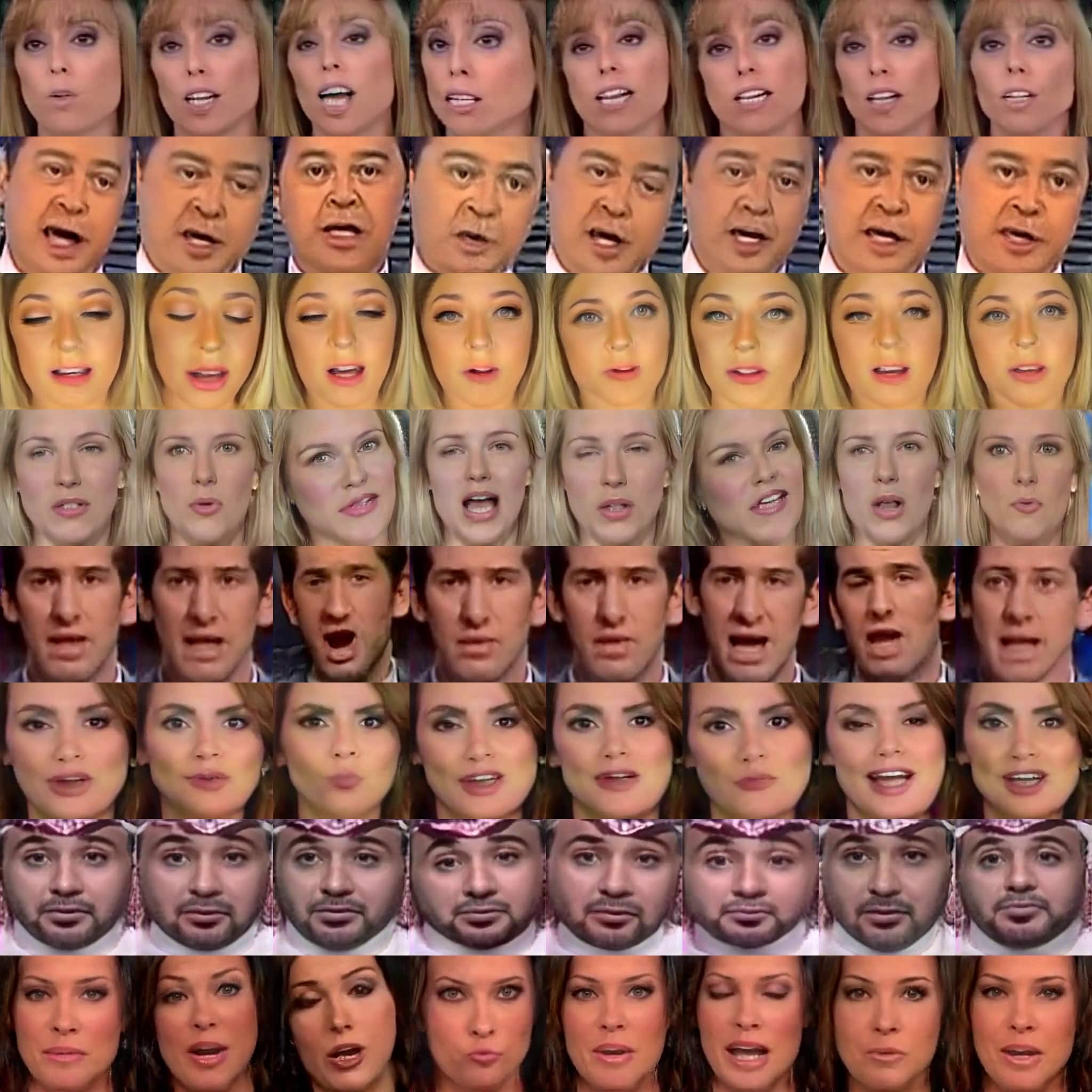}
\caption{Each row is synthesized using the same content code to generate diverse motion patterns. Please see the corresponding supplementary video for a better illustration.}\label{fig:face_16_sameCont_diffMotion}
\end{minipage}\hfill 
\begin{minipage}[t]{.49\textwidth}
\centering
% \makebox[0mm][s]{
% \includegraphics[width=0.8\linewidth]{gif/faceforensics_samemotion_diffcont/1.jpg}}
% \animategraphics[width=0.8\linewidth]{10}{gif/faceforensics_samemotion_diffcont/}{1}{16}
\includegraphics[width=0.8\linewidth]{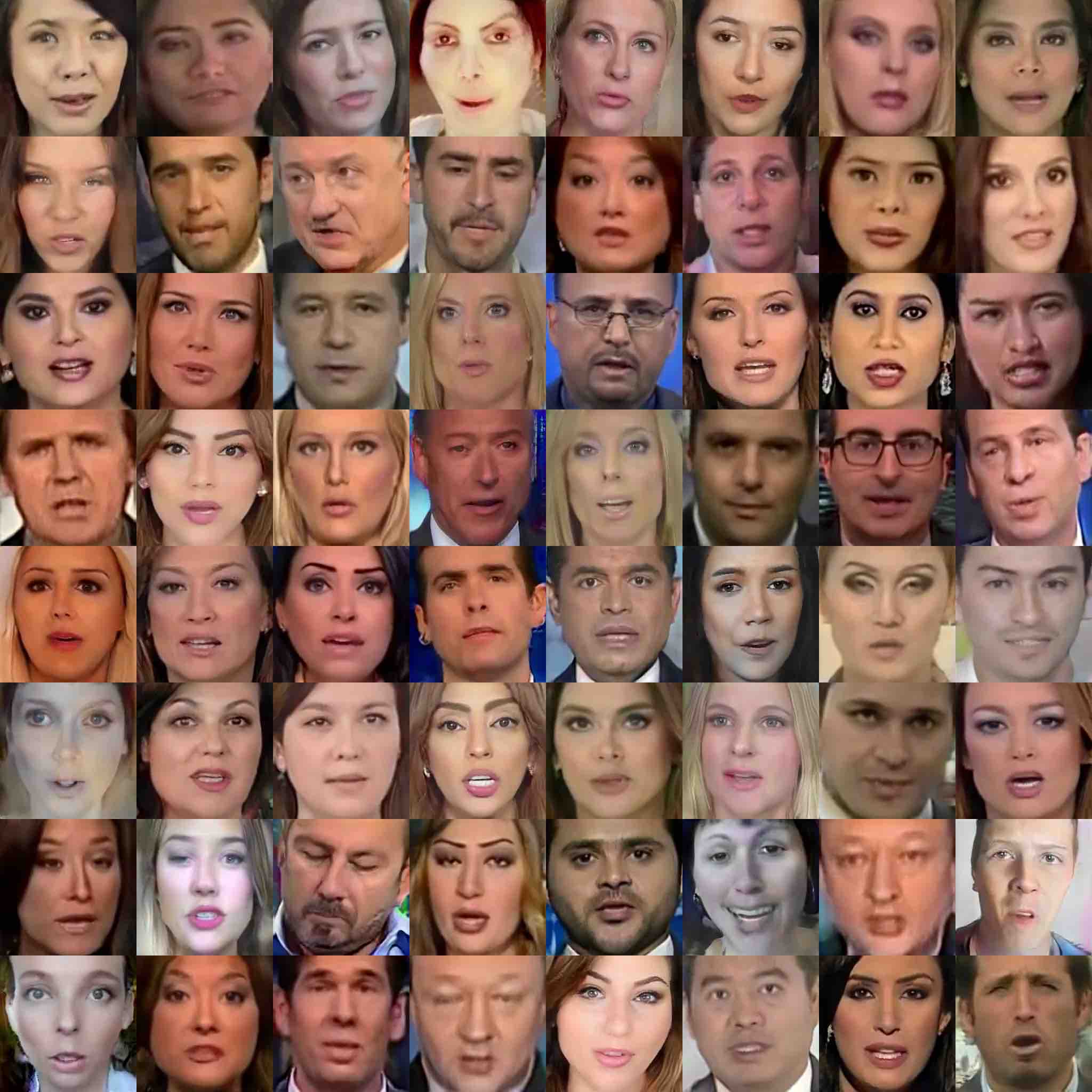}
\caption{Each row is synthesized with the same motion trajectory but different content codes. Please see the corresponding supplementary video for a better illustration.}\label{fig:face_16_sameMotion_diffContent}
\end{minipage}
\end{figure}

\textbf{FaceForensics.} In Fig.~\ref{fig:face_appendix}, we show the generated videos for FaceForensics. 
% We also show the videos synthesized by TGANv2~\citep{TGAN2020} in Fig.~\ref{fig:face_tgan2} for a comparison.
In Fig.~\ref{fig:face_32_appendix} and Fig.~\ref{fig:face_64_appendix}, we show that our approach can generate long consecutive results, $32$ and $64$ frames respectively, even when trained with $16$-frame clips. In Fig.~\ref{fig:face_16_sameCont_diffMotion}, we demonstrate that our approach can generate diverse motion patterns using the same content code. In Fig.~\ref{fig:face_16_sameMotion_diffContent}, we apply the same motion codes with different content to get the synthesized videos.

\begin{figure}[h]
\begin{minipage}[t]{.49\linewidth}
\centering
% \makebox[0mm][s]{
% \includegraphics[width=0.8\linewidth]{gif/sky/1.jpg}}
% \animategraphics[width=0.8\linewidth]{10}{gif/sky/}{1}{16}
\includegraphics[width=0.8\linewidth]{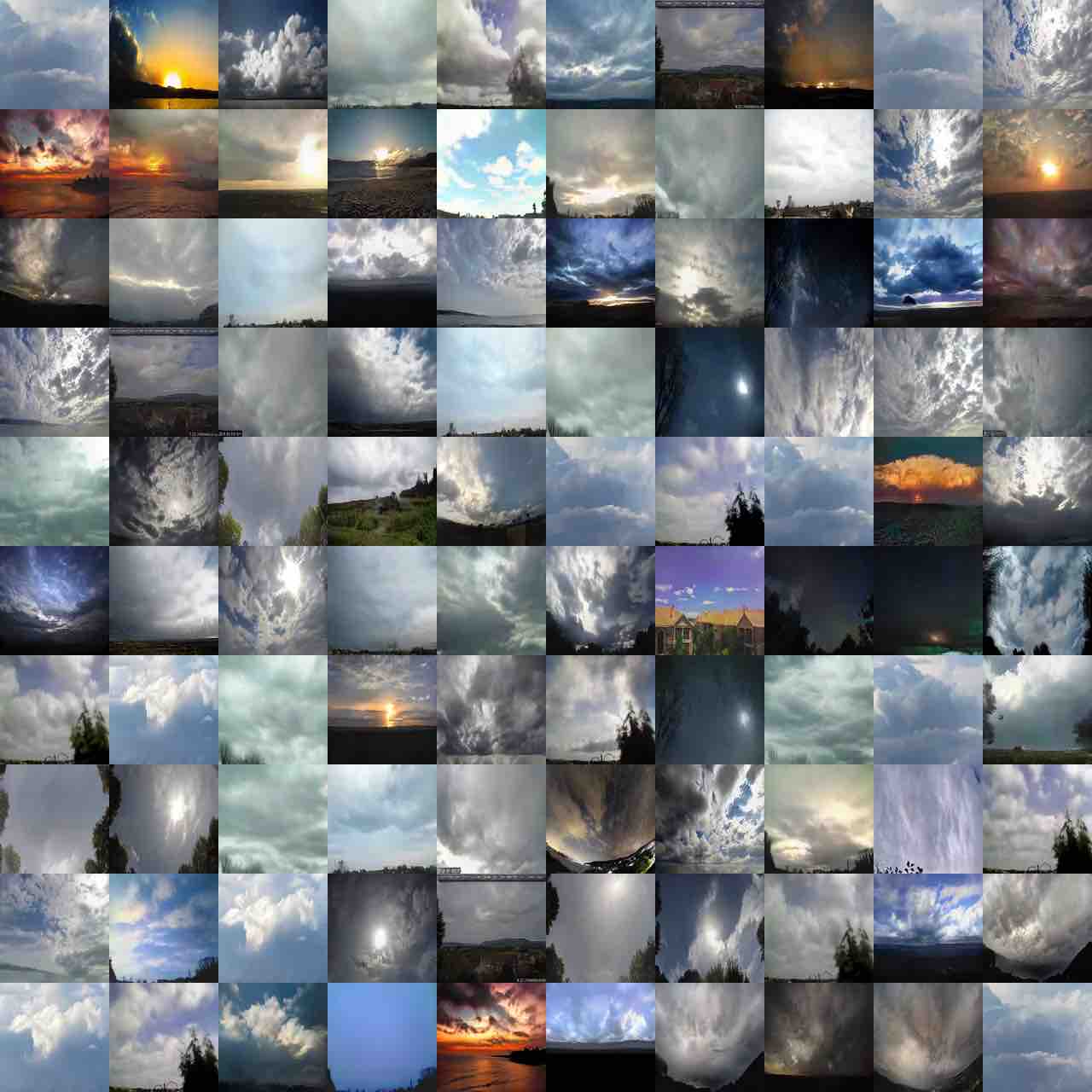}
\caption{Example videos generated by our approach on the Sky Time-lapse dataset. The videos have a resolution of $128\times128$.}\label{fig:sky_app}
\end{minipage}\hfill 
\begin{minipage}[t]{.49\textwidth}
\centering
% \makebox[0mm][s]{
% \includegraphics[width=0.8\linewidth]{gif/ffhq/1.jpg}}
% \animategraphics[width=0.8\linewidth]{10}{gif/ffhq/}{1}{16}
\includegraphics[width=0.8\linewidth]{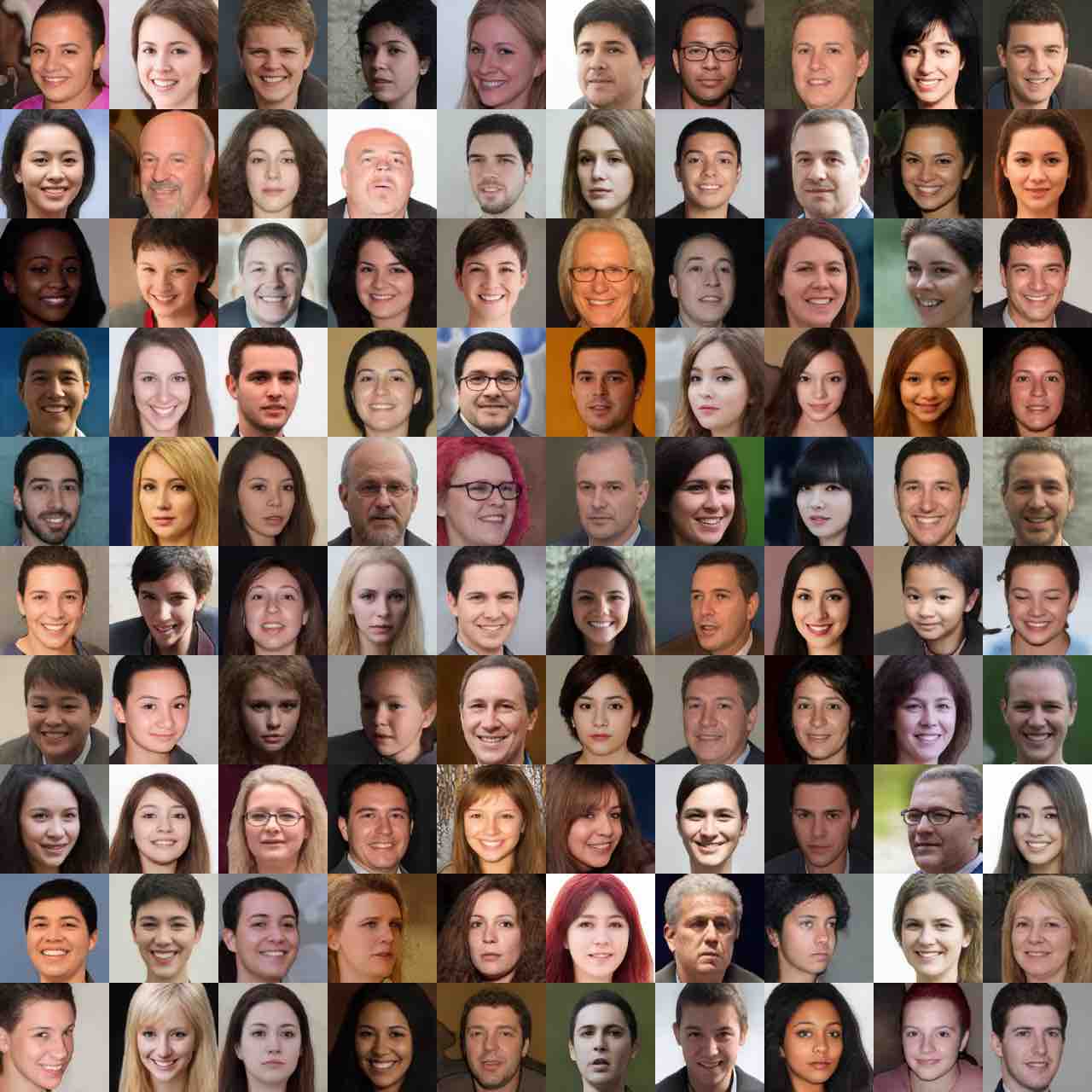}
\caption{Cross-domain video generation for (FFHQ, Vox). The videos have a resolution of $128\times128$.}\label{fig:ffhq-biggan}
\end{minipage}
\end{figure}

\textbf{Sky Time-lapse.} Fig.~\ref{fig:sky_app} shows the generated videos for the Sky Time-lapse dataset. 
% Videos generated by DTVNet~\citep{zhang2020dtvnet} are shown in Fig.~\ref{fig:sky_dtv}.

\begin{figure}[h]
\begin{minipage}[t]{.49\textwidth}
\centering
% \makebox[0mm][s]{
% \includegraphics[width=0.8\linewidth]{gif/ffhq/1.jpg}}
% \animategraphics[width=0.8\linewidth]{10}{gif/ffhq/}{1}{16}
\includegraphics[width=0.8\linewidth]{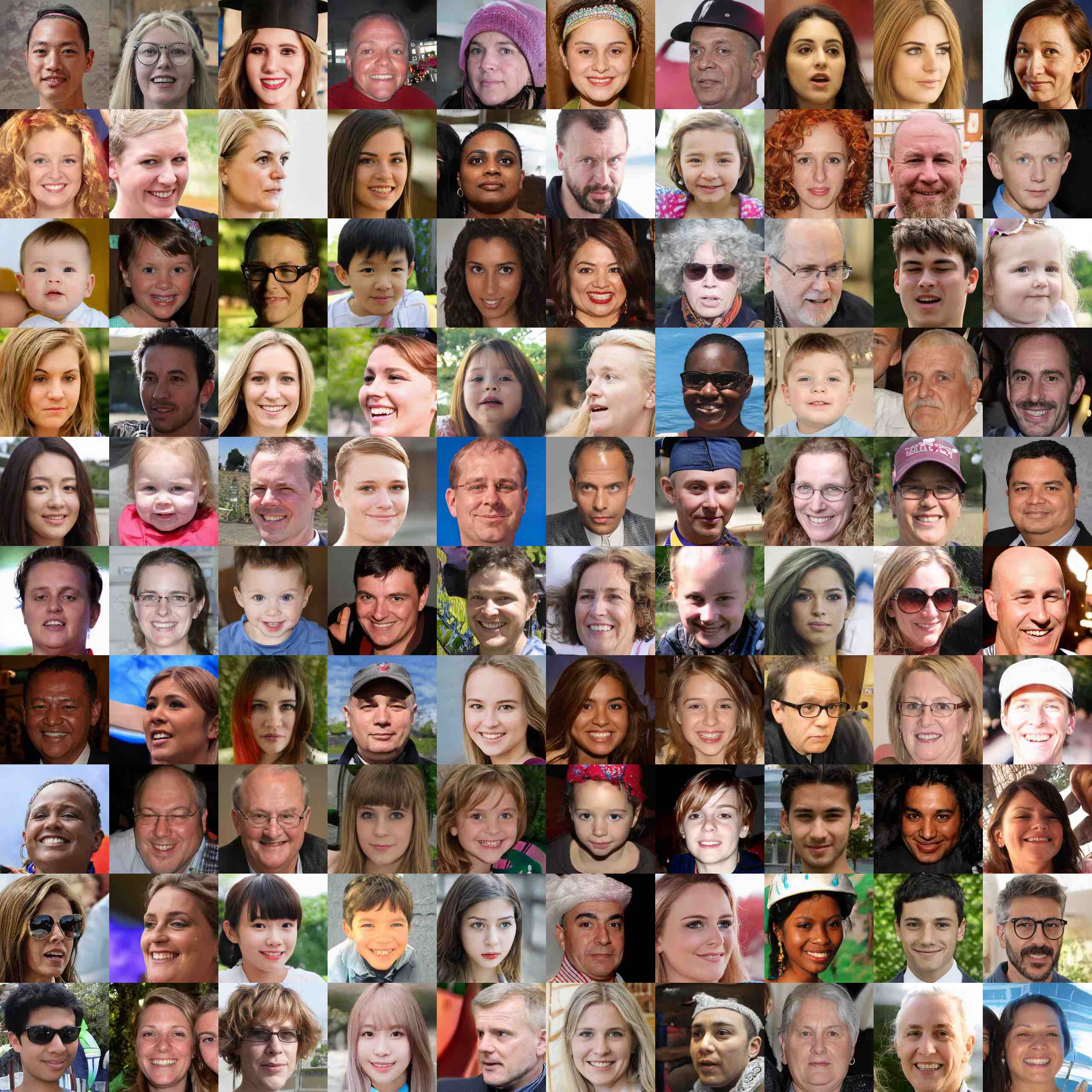}
\caption{Cross-domain video generation for (FFHQ, Vox). The videos have a resolution of $256\times256$.}\label{fig:ffhq-vox}
\end{minipage}\hfill
\begin{minipage}[t]{.49\linewidth}
\centering
% \makebox[0mm][s]{
% \includegraphics[width=0.8\linewidth]{gif/ffhq-1024/1.jpg}}
% \animategraphics[width=0.8\linewidth]{10}{gif/ffhq-1024/}{1}{16}
\includegraphics[width=0.8\linewidth]{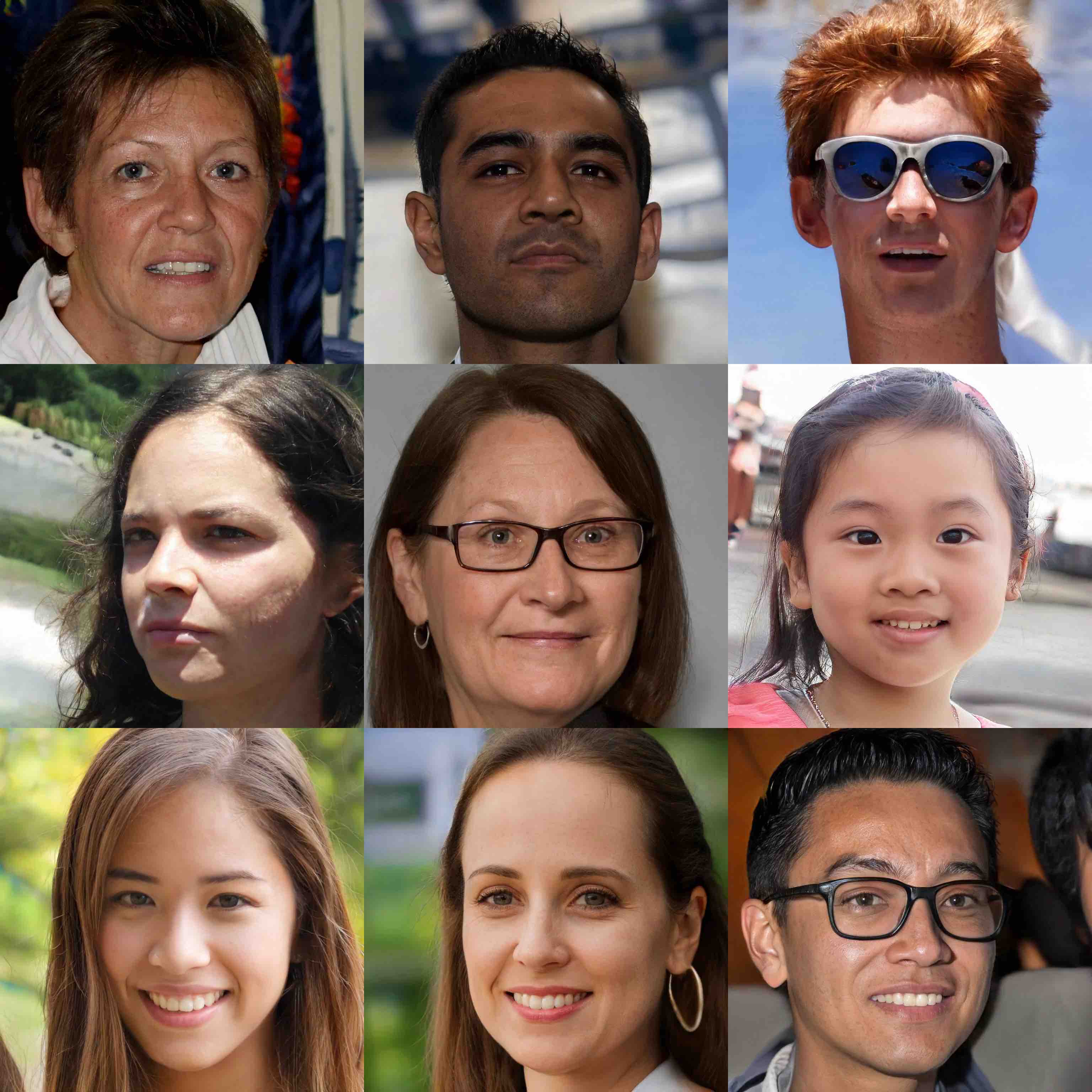}
\caption{Cross-domain video generation for (FFHQ, Vox). The videos have a resolution of $1024\times1024$.}\label{fig:ffhq-vox-1024}
\end{minipage}
\end{figure}

\textbf{(FFHQ, VoxCeleb).} Fig.~\ref{fig:ffhq-biggan}, Fig.~\ref{fig:ffhq-vox}, and Fig.~\ref{fig:ffhq-vox-1024} present the generated videos that have motion patterns from VoxCeleb and content from FFHQ, with resolutions of $128\times128$, $256\times256$, and $1024\times1024$, respectively. We use BigGAN as the generator for Fig.~\ref{fig:ffhq-biggan} and StyleGAN2 for Fig.~\ref{fig:ffhq-vox} and Fig.~\ref{fig:ffhq-vox-1024}.

\begin{figure}[h]
\begin{minipage}[t]{.49\textwidth}
\centering
% \makebox[0mm][s]{
% \includegraphics[width=0.8\linewidth]{gif/afhq_dog/1.jpg}}
% \animategraphics[width=0.8\linewidth]{10}{gif/afhq_dog/}{1}{16}
\includegraphics[width=0.8\linewidth]{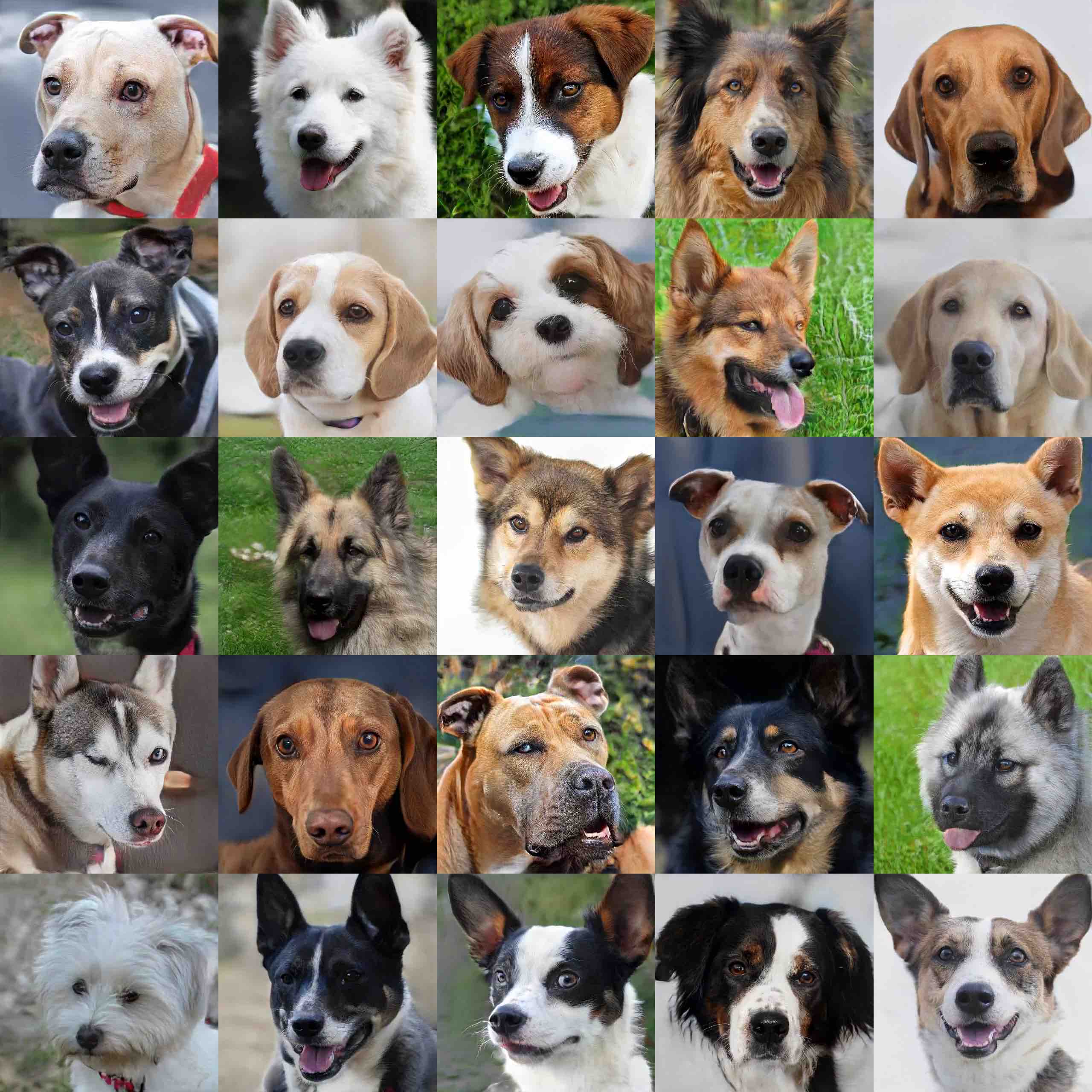}
\caption{
Cross-domain video generation for (AFHQ-Dog, Vox). The videos have a resolution of $512\times512$.}\label{fig:afhq-dog}
\end{minipage}\hfill
\begin{minipage}[t]{.49\linewidth}
\centering
% \makebox[0mm][s]{
% \includegraphics[width=0.8\linewidth]{gif/afhq_dog32/1.jpg}}
% \animategraphics[width=0.8\linewidth]{10}{gif/afhq_dog32/}{1}{16}
\includegraphics[width=0.8\linewidth]{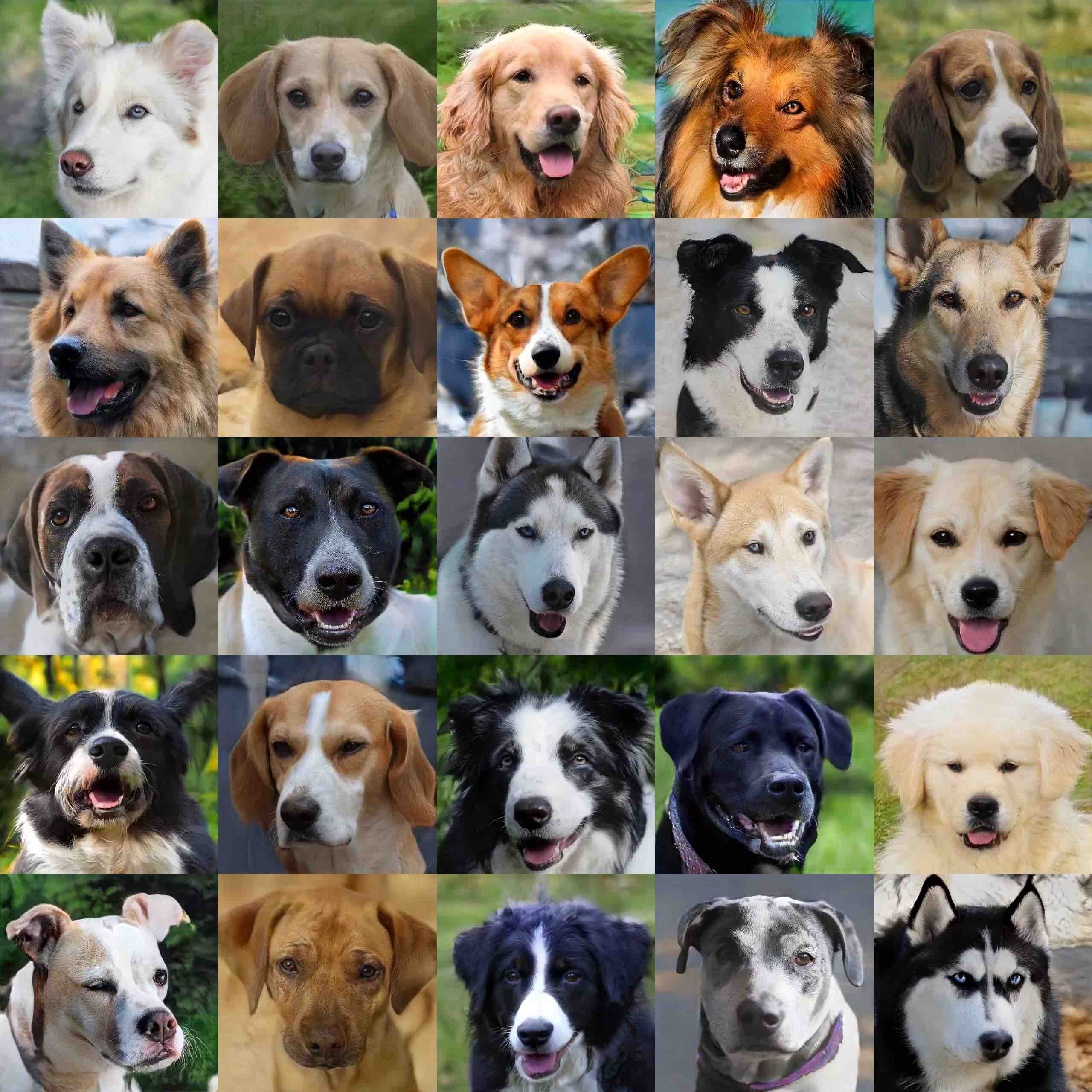}
\caption{Cross-domain video generation for (AFHQ-Dog, Vox). We interpolate every two frames to get 32 sequential frames. The videos have a resolution of $512\times512$.}\label{fig:afhq-dog-interpolate}
\end{minipage}
\end{figure}

\textbf{(AFHQ-Dog, VoxCeleb).} Fig.~\ref{fig:afhq-dog} presents the generated videos that have motion patterns from VoxCeleb and content from AFHQ-Dog. The videos have a resolution of $512\times512$. In Fig.~\ref{fig:afhq-dog-interpolate}, we show the interpolation between every two frames to get longer sequences. 
% in the two figures have the resolution of  resolution in Fig.~\ref{fig:ffhq-vox} is $256\times256$ and $1024\times1024$ in Fig.~\ref{fig:ffhq-vox-1024}.

\begin{figure}[h]
\begin{minipage}[t] {.49\textwidth}
\centering
% \makebox[0mm][s]{
% \includegraphics[width=0.8\linewidth]{gif/anime/1.jpg}}
% \animategraphics[width=0.8\linewidth]{10}{gif/anime/}{1}{16}
\includegraphics[width=0.8\linewidth]{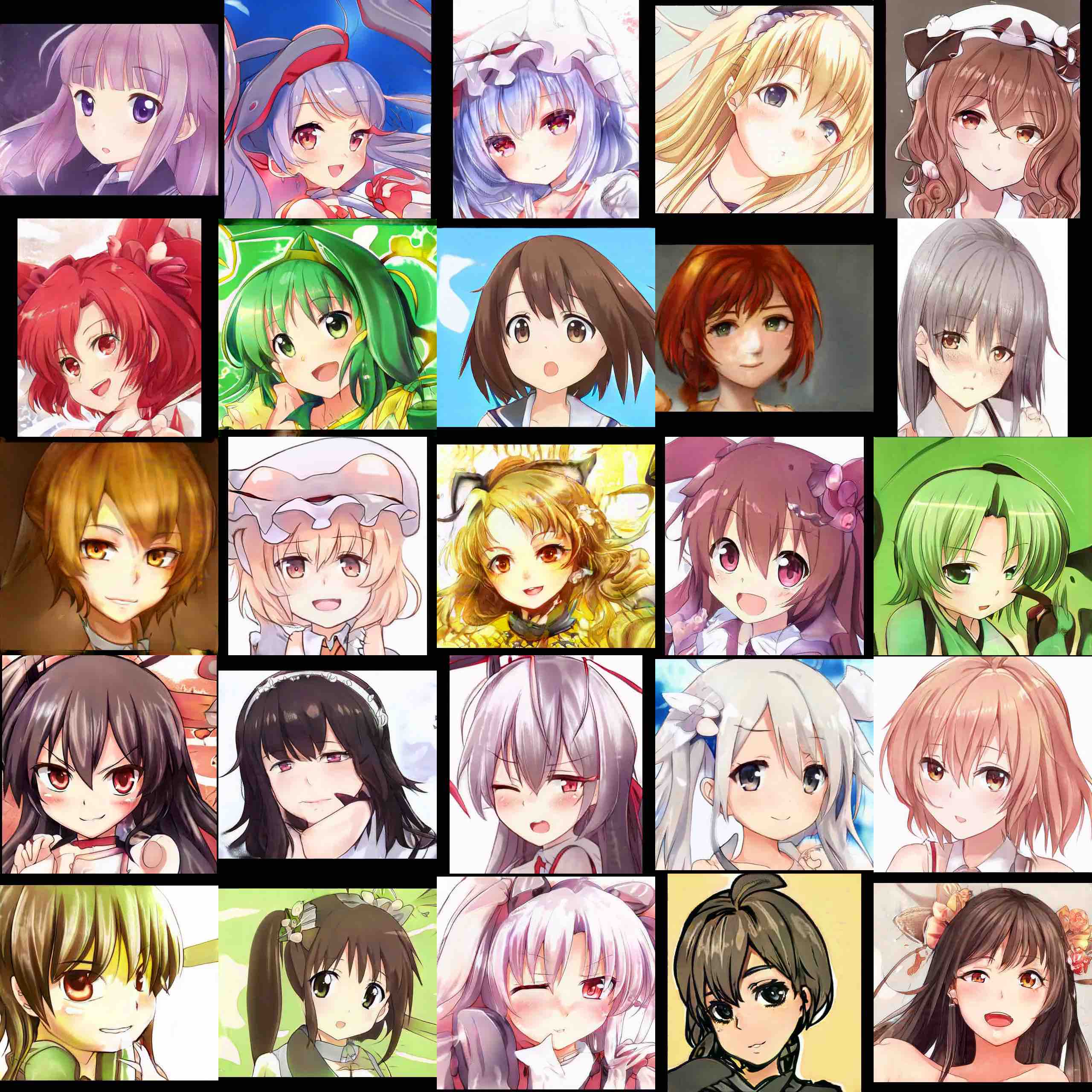}
\caption{Cross-domain video generation for (AnimeFaces, Vox). The videos have a resolution of $512\times512$.}\label{fig:anime-vox}
\end{minipage}\hfill
\begin{minipage}[t]{.49\textwidth}
\centering
% \makebox[0mm][s]{
% \includegraphics[width=0.8\linewidth]{gif/church/1.jpg}}
% \animategraphics[width=0.8\linewidth]{10}{gif/church/}{1}{16}
\includegraphics[width=0.8\linewidth]{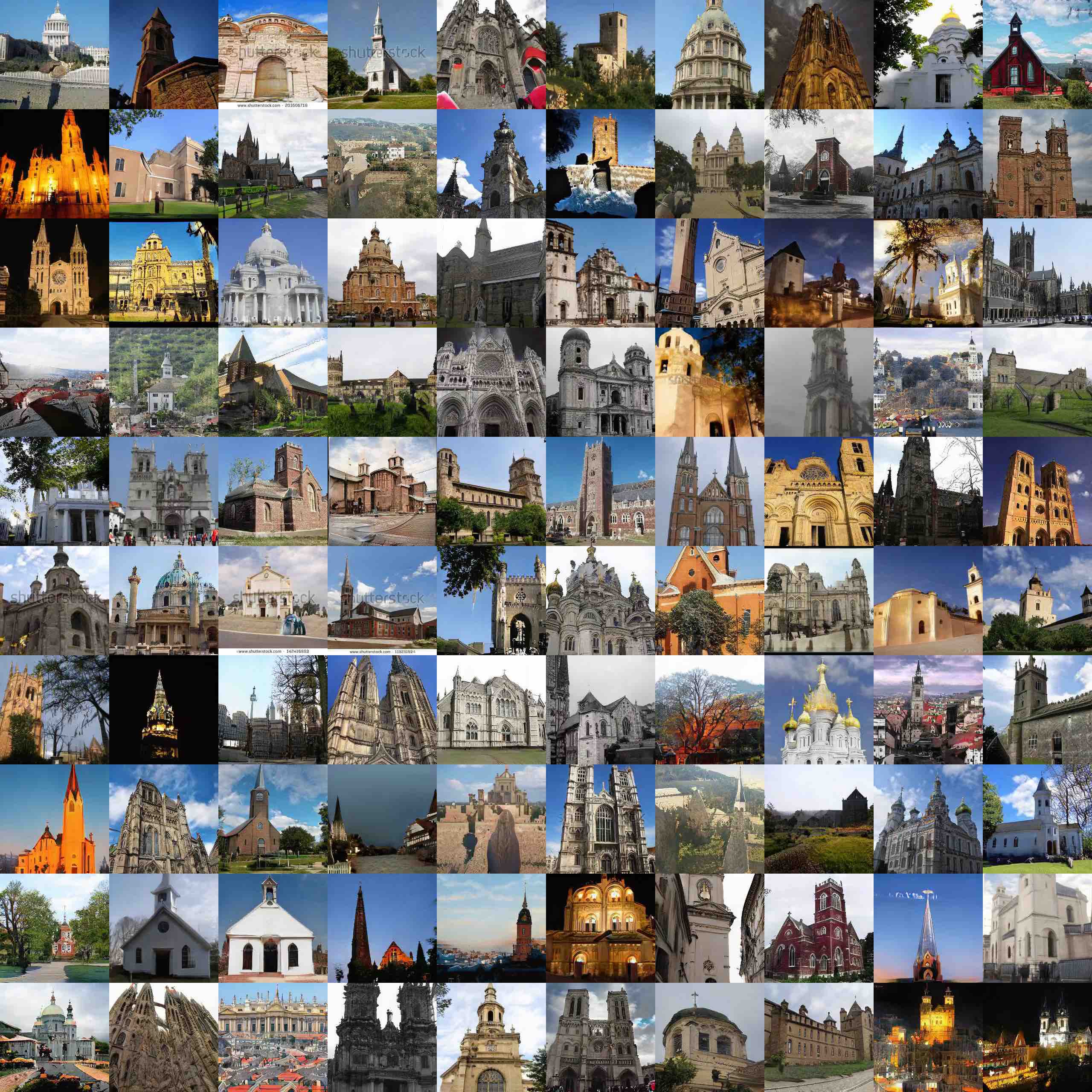}
\caption{Cross-domain video generation for (LSUN-Church, TLVDB). The videos have a resolution of $256\times256$.}\label{fig:church}
\end{minipage}

\end{figure}

\textbf{(AnimeFaces, VoxCeleb).} Fig.~\ref{fig:anime-vox} shows the generated videos that have motion patterns from VoxCeleb and content from AmimeFaces. The videos have a resolution of $512\times512$.

\textbf{(LSUN-Church, TLVDB).} Fig.~\ref{fig:church} presents the generated videos that have time-lapse changing style from TLVDB and content from LSUN-Church.

% \textcolor{red}{Introduce the figures/videos in Appendix.}

% \subsection{More Analysis for Ablation Study}
% \textcolor{blue}{The module ablation results can be found in Table \ref{tab:ablation}. Except w/o PCA, all other models have comparable IS, indicating our residual design makes it easier to model content and motion in $\mathcal{W}$ space. For w/o 3D, surprisingly, it beats prior state-of-the-arts in both IS and FVD. We contribute the reason to our controlled residual estimation and the contrastive loss, where the former one makes our generation has moderate motion change, and the latter one restricts the video content. We also noticed that w/o 3D is worse than w/o 2D in terms of FVD, illustrates the 3D discriminator is important for motion synthesis. Our full model, which has above modules, has the IS and FVD. We additionally train the whole model with a low resolution image generator (128$\times$128), the result show that a better image generator benefits our framework.} 

% \textcolor{red}{All the videos are here \url{https://drive.google.com/drive/folders/1fpzc8-4sbnKgLUbn96_nib0WUZ-8Ljdd?usp=sharing}}

% UCF and FaceForensics
% =============================

%FaceForensics32 & FaceForensics64
% =============================

%  samecontent diffmotion & ameMotion DiffContent
% =============================

% sky & ffhq
% =============================

% ffhq-256 and ffhq-1024
% =============================

% AFHQ_DOG and AFHQ_DOG-32
% =============================

%  anime and church
% =============================
    
% \begin{figure}[h]
% \centering
% \includegraphics[width=1\linewidth]{appendix_unroll/app_ucf.jpg}
% \caption{UCF.}\label{fig:ucf_app}
% \end{figure}

% \begin{figure}[h]
% \centering
% \includegraphics[width=1\linewidth]{appendix_unroll/app_face.jpg}
% \caption{FaceForensics.}\label{fig:face_app}
% \end{figure}

% \begin{figure}[h]
% \centering
% \includegraphics[width=1\linewidth]{appendix_unroll/app_face32.jpg}
% \caption{FaceForensics 32 frames.}\label{fig:face32_app}
% \end{figure}

\newpage 
\cleardoublepage
{\section{More Ablation Analysis For Mutual Information Loss $\mathcal{L}_{\mathrm{m}}$}}
{
In addition to Tab.~\ref{tab:ablation-face}, we perform another ablation experiment to show how mutual information loss $\mathcal{L}_{\mathrm{m}}$ improves motion diversity by considering the following setting. 
We random sample a content code $z_{1}\in\mathcal{Z}$ and use it 
% For a randomly sampled content code $z_{1}\in\mathcal{Z}$, we use it 
as an input to synthesize $100$ videos, where each video contains $16$ frames.
% For each frame of these $100$ videos (they share the same first frame), we get their per-pixel mean and standard deviation (std) of the $100$ frames.
% mean video and per-pixel standard deviations. 
% A blurry mean video and 
We average the generated $100$ videos (they share the same first frame) to get one \emph{mean-video}, which contains $16$ frames.
For example, for the last frame in the mean-video, it is obtained by averaging all the last frames from the $100$ generated videos. 
We also calculate the per-pixel standard deviation (std) for each averaged frame in the mean-video.
More blurry frames and higher per-pixel std indicate the $100$ synthetic videos contain more diverse motion. 
}

\begin{figure}[hbt]
\begin{center}
\includegraphics[width=1\linewidth]{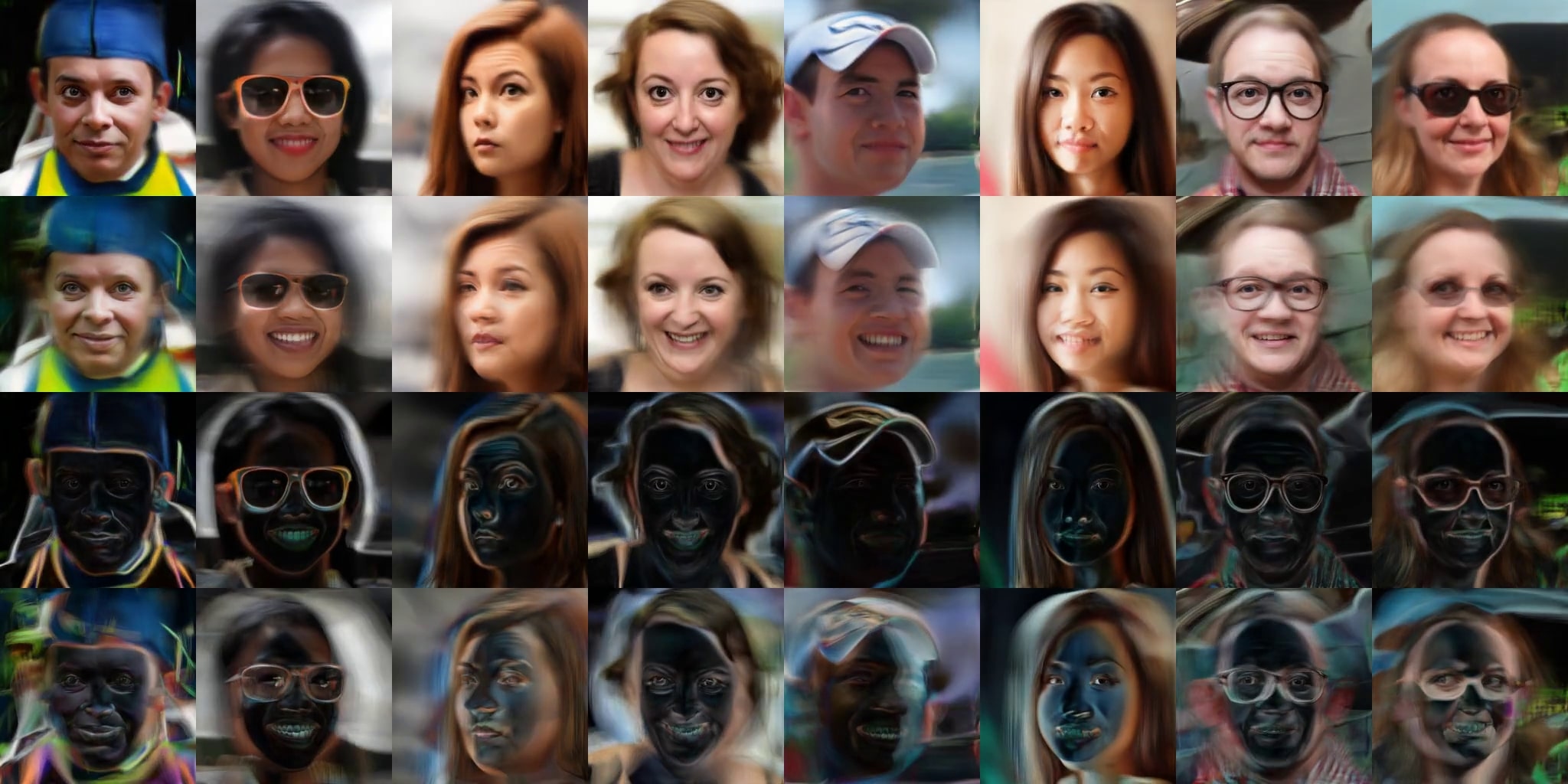}
\caption{{\textbf{Row 1 and 3}: The last frame of the mean-video and per-pixel std of \textit{w/o $\mathcal{L}_\mathrm{m}$} model. \textbf{Row 2 and 4}: The last frame of the mean-video and per-pixel std of the \textit{Full} model. The \textit{Full} model has a more blurry mean-video and higher per-pixel std, which indicates more diverse motion.}}\label{fig:diverse-lm}
\end{center}
\end{figure}

{
We evaluate the settings of \textit{Full} and \textit{w/o $\mathcal{L}_\mathrm{m}$} (without using the mutual information loss) by running the above experiments for $50$ times, \textit{e.g.}, sampling $z_{1}$ for $50$ times. Across the $50$ trials, for \textit{Full} model, the mean and std of the per-pixel std for the $16^{th}$ frame (the \emph{last} frame in a generated video) is $0.233\pm 0.036$, which is significantly higher than that of the \textit{w/o $\mathcal{L}_\mathrm{m}$} model ($0.126\pm 0.025$). In Fig.~\ref{fig:diverse-lm}, we show $8$ examples of the last frame from the mean-video and the images with per-pixel std (See supplementary material for the whole videos). Our \textit{Full} model has more diverse motion as the averaged frame is more blurry and the per-pixel std is higher. Note that StyleGAN2 enables noise inputs for extra randomness, we disable it in this ablation study.}

{\section{Limitations}}\label{app:limitations}
{
Our framework requires a well-trained image generator for frame synthesis. In order to synthesize high-quality and temporally coherent videos, an \emph{ideal} image generator should satisfy two requirements: {\bf R1.} The image generator should synthesize high-quality images, otherwise the video discriminator can easily tell the generated videos as the image quality is different from the real videos. {\bf R2.} The image generator should be able to generate diverse image contents to include enough motion modes for sequence modeling.}

{\bf Example of R1.} UCF-101 is a challenging dataset even for the training of an \emph{image} generator. In Tab.~\ref{tab:g-fid}, the StyleGAN2 model trained on UCF-101 has FID $45.63$, which is much worse than the others. We hypothesis the reason is that {UCF-101 dataset has many categories, but within each category, it} includes relatively a small amount of videos and these videos share very similar content. 
% which make the training of image generator , and the intra-class diversity can be restricted to just a handful of videos per class, 
Such observation is also discussed in DVDGAN~\citep{clark2019adversarial}. Although we can achieve state-of-the-art performance on UCF-101 dataset, the quality of the generated videos is not as good as other datasets (Fig.~\ref{fig:ucf_qualitative}), and the quality of synthesized videos is still not close to real videos.

{
{\bf Example of R2.}
We test our method on BAIR Robot Pushing Dataset~\citep{ebert2017self}. We train a $64\times 64$ StyleGAN2 image generator with using the frames from BAIR videos. The image generator has FID as $6.12$. Based on the image generator, we train a video generation model that can synthesize $16$ frames. An example of synthesized video is shown in Fig.~\ref{fig:bair_example} (more videos are in the supplementary materials). We can see our method can successfully model shadow changing, the robot arm moving, but it struggles to decouple the robot arm from some small objects in the background, which we show analysis follows.}

\begin{figure}[hbt]
\begin{center}
\includegraphics[width=1\linewidth]{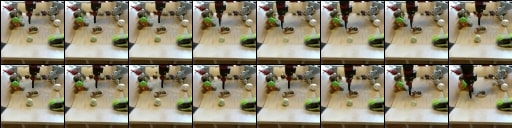}
\caption{{A synthesized video using BAIR dataset. Note the background changing of the first frame (upper-left) and the last frame (bottom-right).}}\label{fig:bair_example}
\end{center}
\end{figure}

\subsection{{Analysis of the information contained in PCA components.}}
\begin{figure}[hbt]
\begin{center}
\includegraphics[width=0.9\linewidth]{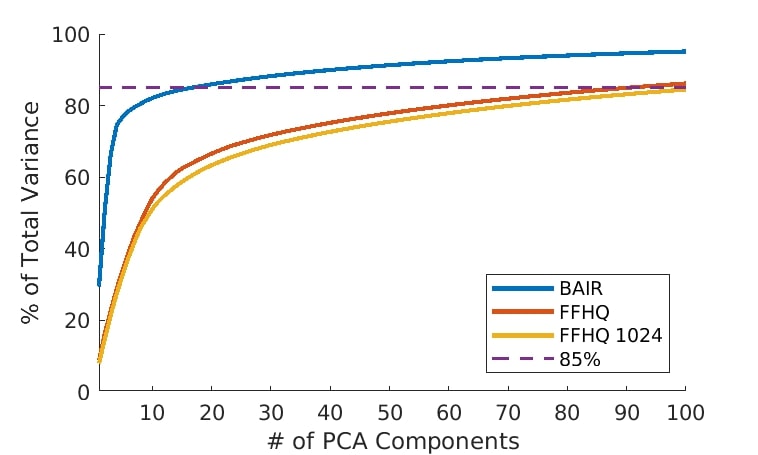}
\caption{{Percentage of variations captured by top PCA components on different models.}}\label{fig:var_percent}
\end{center}
\end{figure}

{
Inspired by previous work~\citep{harkonen2020ganspace}, we further investigate the latent space of the image generator by considering the information contained in each PCA component.
% how many PCA components are needed in the input latent space of the image generator.
% , that is trained on BAIR. 
Fig.~\ref{fig:var_percent} shows the percentage of total variance captured by top PCA components. 
% Compared to top PCA components on FFHQ dataset, 
The image generator on BAIR compresses most of the information on a \emph{few} components. Specially, the top $20$ PCA components captures $85\%$ of the variance. 
In contrast, the latent space of the image generator trained on FFHQ (and FFHQ 1024 for high-resolution image synthesis) uses $100$ PCA components to capture $85\%$ information. 
This implies the BAIR generator models the dataset in a low-dimension space, and such generator increases the difficulty for fully disentangling all the objects  in images for manipulation.}

\begin{figure}[h]
\begin{minipage}[t] {.49\textwidth}
\centering
\includegraphics[width=0.8\linewidth]{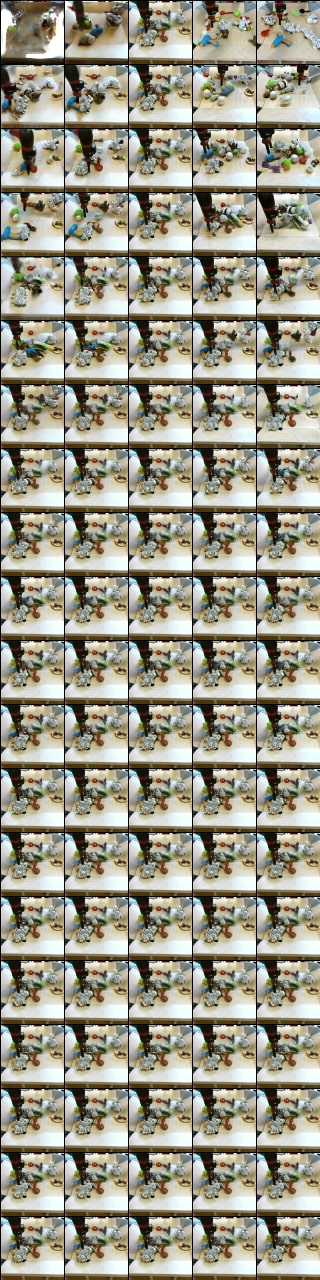}
\end{minipage}\hfill
\begin{minipage}[t]{.49\textwidth}
\centering
\includegraphics[width=0.8\linewidth]{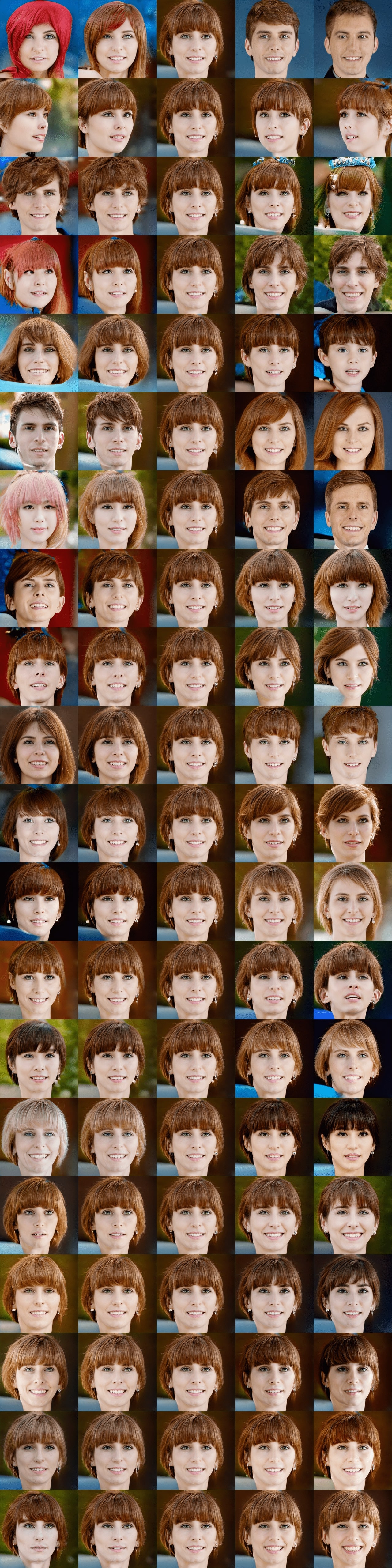}
\end{minipage}
\caption{{Visualization of top $20$ principle components of BAIR (left) and FFHQ (right).}}\label{fig:20-pca}
\end{figure}

{
Moreover, we visualize the video synthesis results by moving along the top $20$ PCA components. 
Let $V_{i}$ denote the $i^{th}$ PCA component. 
Given content code $z_{1}$, we synthesize a $5$-frame video clip by using the following sequence as input: $\{z_{1}-2V_{i}, z_{1}-V_{i}, z_{1}, z_{1}+V_{i}, z_{1}+2V_{i}\}$. 
In Fig.~\ref{fig:20-pca}, we show the video synthesis results by moving along the top $20$ PCA directions. It can be seen that: 1) changing the later components (the $8^{th}$ and later rows) of BAIR only make small changes; 2) the first $7$ components of BAIR have entangled semantic meaning, while the components in FFHQ have more disentangled meaning ($2^{nd}$ row, rotation; $20^{th}$ row, smile). This indicates the image generator of BAIR may not cover enough (disentangled) motion modes, and it might be hard for the motion generator to fully disentangle all the contents and motion with only a few dominating PCA components, while for the image generator trained on FFHQ, it is much easier for disentangling foreground and background.}

% \section{\textcolor{blue}{Future Work}}
% In future work, instead of \textcolor{red}{training image generator separately}, we plan to improve the training of image generator by incorporating video-related supervisions. For example, frames within same video clip should have similar content and different motion, and such supervision should improve the semantic behavior of the latent space of StyleGAN2/BigGAN.
\end{document}